\documentclass{article}

\usepackage[numbers, sort, comma, square]{natbib}

\usepackage[final]{neurips_2021}

\usepackage[utf8]{inputenc} %
\usepackage[T1]{fontenc}    %
\usepackage{hyperref}       %
\usepackage{url}            %
\usepackage{booktabs}       %
\usepackage{amsfonts}       %
\usepackage{nicefrac}       %
\usepackage{microtype}      %
\usepackage{xcolor}         %
\usepackage{graphicx}
\usepackage[english]{babel}
\usepackage[font=small,labelfont=bf]{caption}
\usepackage{multirow}
\usepackage{amsmath}
\usepackage{multicol,lipsum}
\usepackage{array}
\usepackage{tabu}
\usepackage{epsfig}
\usepackage{parallel}
\usepackage{diagbox}
\usepackage{mathtools}
\usepackage{wrapfig}
\usepackage{makecell}
\usepackage{algorithm}
\usepackage{algorithmic}
\usepackage{caption} 
\usepackage{subcaption}
\usepackage{mwe}
\newcommand{\Secref}[1]{Section~\ref{#1}}
\newcommand{\secref}[1]{Section~\ref{#1}}

\newcommand{\Figref}[1]{Figure~\ref{#1}}

\newcommand{\Tabref}[1]{Table~\ref{#1}}

\newcommand{\Appref}[1]{Appendix~\ref{#1}}

\renewcommand{\eqref}[1]{Eqn.~\ref{#1}}
\newcommand{\Algref}[1]{Alg.~\ref{#1}}

\definecolor{ourRed}{HTML}{E24A33}
\definecolor{ourBlue}{HTML}{348ABD}
\definecolor{ourPurple}{HTML}{988ED5}
\definecolor{ourGray}{HTML}{777777}
\definecolor{ourLightGray}{HTML}{B8B8B8}
\definecolor{ourYellow}{HTML}{FBC15E}
\definecolor{ourGreen}{HTML}{4D8951}
\definecolor{ourPink}{HTML}{FFB5B8}
\definecolor{oursteelblue}{HTML}{9BB8D7}
\definecolor{ourOrange}{HTML}{FDBA58}
\definecolor{ourWhite}{HTML}{FAFAFA}

\newcommand{\actionSeq}[1]{\langle#1\rangle}

\newcommand{\contribfootnote}[1]{%
  \begingroup
  \renewcommand{\thefootnote}{}\footnote{#1}%
  \addtocounter{footnote}{-1}%
  \endgroup
}

\newcommand{\PaperTitle}{%
ReaSCAN: Compositional Reasoning in\\ Language Grounding%
}

\title{\PaperTitle}

\author{%
 Zhengxuan Wu$^{\ast}$ \\
  Stanford University \\
  \texttt{\href{mailto:wuzhengx@stanford.edu}{wuzhengx@stanford.edu}} \\
  \And
  Elisa Kreiss$^{\ast}$  \\
  Stanford University \\
  \texttt{\href{mailto:ekreiss@stanford.edu}{ekreiss@stanford.edu}} \\
  \AND
  Desmond C.~Ong \\
  National University of Singapore \\
  \& IHPC, A*STAR \\ %
  \texttt{\href{mailto:dco@comp.nus.edu.sg}{dco@comp.nus.edu.sg}} \\
  \And
  Christopher Potts \\
  Stanford University \\
  \texttt{\href{mailto:cgpotts@stanford.edu}{cgpotts@stanford.edu}}
}

\begin{document}

\maketitle

\begin{abstract}
The ability to compositionally map language to referents, relations, and actions is an essential component of language understanding. The recent gSCAN dataset (Ruis et al.~2020, \emph{NeurIPS}) is an inspiring attempt to assess the capacity of models to learn this kind of grounding in scenarios involving navigational instructions. However, we show that gSCAN's highly constrained design means that it does not require compositional interpretation and that many details of its instructions and scenarios are not required for task success. To address these limitations, we propose \textbf{ReaSCAN}, a benchmark dataset that builds off gSCAN but requires compositional language interpretation and reasoning about entities and relations. We assess two models on ReaSCAN: a multi-modal baseline and a state-of-the-art graph convolutional neural model. These experiments show that ReaSCAN is substantially harder than gSCAN for both neural architectures. This suggests that ReaSCAN can serve as a valuable benchmark for advancing our understanding of models' compositional generalization and reasoning capabilities.\contribfootnote{$^{\ast}$Equal contribution.}
\end{abstract}

\section{Introduction}

Natural languages are \emph{compositional} \cite{Montague74,Partee84,Janssen97} and \emph{grounded} \citep{Clark96,Parikh01,harnad1990symbol}; the meanings of complex phrases are derived from their parts, and meaning itself is defined by a mapping from language to referents, relations, and actions. It is therefore vital that we push NLP systems to be grounded and compositional as well. However, the major benchmarks in the field right now mostly do not support rich grounding, and it is often unclear whether they support learning compositional structures, as evidenced by their common failures at simple adversarial tests involving compositionality \cite{glockner-etal-2018-breaking,naik-etal-2018-stress,nie2019analyzing}.

There are several benchmarks for testing compositional generalization~\cite{johnson2017clevr, hill2018understanding, lake2018generalization, hill2020environmental, ruis2020benchmark}. SCAN~\cite{lake2018generalization} focuses on compositionality in the area of interpreting navigational instructions. Building off SCAN, \citet{ruis2020benchmark} propose a grounded version of SCAN called gSCAN, in which agents have to ground navigation commands in a grid world in order to identify the correct referent. gSCAN supports learning in idealized scenarios involving navigational instructions, and it seeks to probe for compositionality. The design is simple and flexible, making it a potentially valuable benchmark and a source for insights into how to design robust tests of language understanding.

However, we find that gSCAN has three major limitations: (1) its set of instructions is so constrained that preserving the linguistic structure of the command is not required; (2) the distractor objects in its grounded scenarios are mostly not relevant for accurate understanding; and (3) in many examples, not all modifiers in the command are required for successful navigation, which further erodes the need for compositional interpretation and inflates model performance scores. 

\begin{figure}[t!]
\centering
     \includegraphics[width=1.0\textwidth]{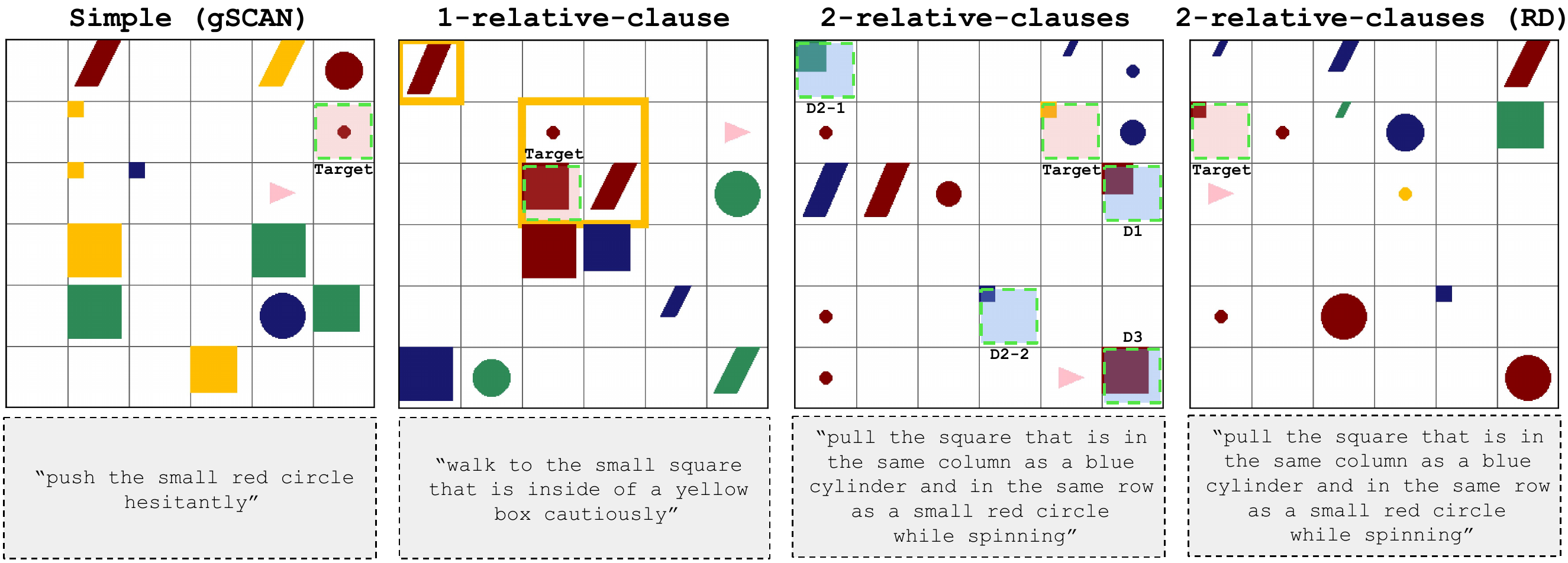}
      \caption{Four command--world pairs for different command patterns. ReaSCAN's simple command is equivalent to gSCAN commands~\cite{ruis2020benchmark} but the structure of the sampled grid world differs. \texttt{RD} indicates that distractors are randomly sampled. A sample of potential referent targets is highlighted with green dashes. The actual target for the given the command is shaded in red and its direct distractors are shaded in blue.}
       \label{fig:reascan-examples}
\end{figure}

In this paper, we propose \textbf{ReaSCAN}, a benchmark dataset that builds off gSCAN and addresses its limitations. \Figref{fig:reascan-examples} provides examples and a comparison with gSCAN. We establish that ReaSCAN requires both compositional language interpretation and complex reasoning about entities and relations. Like gSCAN, ReaSCAN is algorithmically generated, which allows us to vary the difficulty of the learning problems we pose and thus diagnose model limitations with precision. In addition, we introduce a range of complex distractor sampling strategies which, in case of incorrect target identification, can help pinpoint which failure in command understanding led to the error. This allows us to show that challenging distractors can severely impact performance in this task.

We assess two models on ReaSCAN: a multi-modal baseline and a state-of-the-art graph convolutional neural model. These experiments show that ReaSCAN is substantially harder than gSCAN for both neural architectures, and they verify that we can modify the difficulty of learning tasks in the desired ways to achieve fine-grained insights into model performance and model limitations. This suggests that ReaSCAN can serve as a valuable benchmark for advancing our understanding of models' compositional generalization and reasoning capabilities in linguistic tasks. We hope also that the general techniques used to move from gSCAN to ReaSCAN can be applied more generally in the design of future benchmarks for assessing grounded, compositional language use.

\section{Related Work}

There are a variety of efforts underway to more deeply understand how neural models ground linguistic cues with visual inputs, including visual question answering~\cite{johnson2017clevr, yi2019clevrer, vedantam2020curi, suhr2017corpus, hudson2019gqa}, image captioning~\cite{mao2016generation, vedantam2017context, hong2019learning}, referring expression resolution~\cite{lake2018generalization, ruis2020benchmark}, navigation~\cite{anderson2018vision, zhu2020vision, thomason:corl19} and program induction and synthesis~\cite{zhang2019raven, ellis2020dreamcoder, nie2020bongard}. Similar to previous synthetic benchmarks, our work aims to provide a controlled environment that can be used to evaluate a neural model's generalization capabilities according to a variety of specific generalization tasks. We focus on evaluating compositional generalization with grounded referring expression resolution. 

A number of recent approaches involve generating synthetic datasets to evaluate compositional generalization of neural models~\cite{bowman2015recursive, johnson2017inferring, johnson2017clevr, hill2018understanding, hill2020environmental, chevalier2018babyai, bahdanau2019closure, bahdanau2018systematic}. For instance, \cite{bahdanau2019closure} proposed CLOSURE, a set of unseen testing splits for the CLEVR dataset \cite{johnson2017clevr} which contains synthetically generated 
natural-looking questions about 3D geometric objects. Our work investigates a similar generalization over grounded linguistic inputs in a visual scene but focuses specifically on a model's capability to resolve linguistic compositionality. We evaluate the generalization capabilities of neural models by testing them against unseen compositions of the language input which require grounding in simulated shape worlds.  

Models performing well on gSCAN are a promising first test case for ReaSCAN. Numerous approaches have been proposed to handle at least some of the challenges posed by SCAN and gSCAN datasets including novel data augmentation methods and neural architectures~\cite{bastings2018jump, andreas2019good, lake2019compositional, nye2020learning, chen2020compositional, heinze2020think, gao2020systematic, csordas2021devil}. Successful neural models on gSCAN involve compositional neural networks which increase generalizability~\cite{heinze2020think} and language conditioned graph neural networks for encoding objects~\cite{gao2020systematic}. While these techniques solve some of the simpler splits in gSCAN involving generalization of novel object attributes~\cite{gao2020systematic}, we show that they are still ineffective for similar splits of ReaSCAN in \secref{sec:comp-split-novel-obj-attr}. ReaSCAN, therefore, provides a more challenging benchmark, revealing clear shortcomings of current models' generalization capabilities.

\section{Background: The Grounded SCAN Benchmark (gSCAN)}\label{sec:gscan}

The gSCAN benchmark is an extension of the SCAN dataset \cite{lake2018generalization} with a focus on grounding actions in a changeable environment. In gSCAN, a grid world containing an agent and several shapes is paired with a command, such as ``walk to the red square cautiously''. The goal is to generate an action sequence like $\actionSeq{\texttt{left,right,right,left}}$ that lets the agent execute the command in that particular world to reach the referent target. Adverbs like ``cautiously'' assign specific modes of movement to the overall sequence. gSCAN enables tests for compositional generalization by presenting the model with unseen verb--adverb combinations (``walk cautiously'' vs.\ ``push cautiously''), unseen adjective--noun compositions, unseen color--shape feature co-occurrences on objects, and unseen locations for the target referent.

The guiding ideas behind gSCAN seem powerful and relevant, but we identify four ways in which specific design choices reduce the potential of the dataset to achieve its central goals:

\paragraph{1.~Irrelevance of Word Order}
Since gSCAN is meant to be a simple synthetic dataset, all commands consist of a verb, a noun phrase consisting of a noun with a potential color and/or size modifier, and an optional adverb. Given this template, the word order of the input command is irrelevant for determining the correct action sequence. The words ``walk to the red square cautiously'' can be scrambled and still yield a unique order with only a single potential referent. Consequently, Bag-of-Words accounts are in principle sufficient for encoding the gSCAN commands. As a point of contrast, the commands in the earlier SCAN dataset, such as ``walk twice and jump thrice'', cannot be scrambled in this way without a task-relevant loss of information, and are therefore much more challenging to solve on the command level.

\paragraph{2.~A Limited Test for Linguistic Compositionality}
gSCAN includes a test set in which all commands involve a previously unseen referring expression combination (the novel NP ``yellow square''), with the goal of seeing whether models can predict the meaning of the whole from its parts ``yellow'' and ``square'', which are seen in training. This is a clear test for compositionality. However, unfortunately, the split creation process didn't inherently require an understanding of ``yellow'' and ``square'' to be necessary for a unique identification in a specific world. In the split provided by the authors,\footnote{\url{https://github.com/LauraRuis/groundedSCAN}} both the color and shape feature are only required in 62.7\% of all test examples. (Color is sufficient in 25.2\% of all test examples, shape in 10.6\%, and either of the two in 1.4\% of all cases.)
gSCAN also includes a test split designed to require feature attribute composition: in training, the referent target is never an object with the color feature \emph{red} and shape feature \emph{square}. At test time, only red squares are targets and are referred to with all valid referring expressions (i.e., ``(small|big)? red? square''). As in the previous split, color and shape feature in the command are necessary in just 62.5\% of all test examples, making it equally unsuitable for investigating linguistic compositionality.

\paragraph{3.~Biased Distractor Sampling}
Distractor sampling in gSCAN relies on random selection of all objects that are not mentioned in the command. In general, if the utterance mentions a blue circle, the algorithm creates all possible objects that aren't blue circles. Then, it selects half of them as distractors. There is one exception: if the utterance contains a size modifier (as in ``small blue circle''), there will be a big blue circle as a distractor. Due to the distractor sampling design, simple utterances such as ``the circle'' will only have one distractor, while more complex utterances will have many more. This makes by-chance accuracy dependent on the informativity and complexity of the linguistic expression. 

\paragraph{4.~Too Few Effective Distractors}
As shown in the first example for gSCAN of \Figref{fig:reascan-examples}, the output action sequence stays the same even if we randomly reorder all objects except the referent target. In fact, as long as they don't introduce reference ambiguity, the size, color, and shape of other objects can be modified without any effect on the output action sequence. As a result, grounding is based on essentially two objects, the two red circles. (See \Secref{sec:distractor-sampling} for details about how distractors affect performance.)

In sum, gSCAN provides novel systematic ways of investigating grounded language understanding but it lacks a way to keep investigating the syntactic compositional questions in the command which motivated SCAN. ReaSCAN introduces a more complex command structure that enforces models to retain some linguistic structure to solve it, and contains compositional splits that ensure the necessity of compositional generalization capabilities for the input command. Due to the more complex command structure, this requires elaborate distractor sampling strategies with the goal to make distractors maximally competitive to promote grounding to multiple objects in the world.

\section{The Reasoning-based SCAN Benchmark (ReaSCAN)}\label{sec:reascan}

We now introduce ReaSCAN, which seeks to address the above limitations of gSCAN. Like gSCAN, ReaSCAN is a command-based navigation task that is grounded in a grid world containing an agent, a referent target, and a set of distractors, as shown in \Figref{fig:reascan-examples}. 

Given a command $\textbf{C}_{i}$ paired with a corresponding grid world $\mathcal{W}_{i,j}$, the goal is to generate an action sequence $\textbf{a}_{i,j}$ which contains the actions that the agent needs to take in order to reach the target referent and operate on it.
An oracle model learns a mapping $\mathcal{G}$ that formulates $\textbf{a}_{i,j} = \mathcal{G}(\mathcal{W}_{i,j}, \textbf{C}_{i})$ for $i \in [1, N]$, where $N$ is the number of commands, and $j \in [1, M]$, where $M$ is the number of worlds generated for each command. 

Crucially, ReaSCAN extends gSCAN while ensuring two main desiderata: (1) word-order permutations in the command will lead to ambiguities about the intended referent, requiring a model to resolve linguistic structure, and (2) the identity of the referent depends on reasoning about multiple distractor objects in the world. Consider the \texttt{2-relative-clauses} example (third from left) in \Figref{fig:reascan-examples}. If we scramble the word order of the command by swapping attributes between the second and the third objects, and change them to ``small blue circle'' and ``red cylinder'', the referent target changes (e.g., object \texttt{D1} in the world); additionally, if the model only understands the first relational clause ``the same column as a blue cylinder'', it may discover multiple referent targets (e.g., object \texttt{D2-1} in the world). These modifications ensure that understanding ReaSCAN commands requires resolving the syntactic structure of the command, while largely maintaining the simplicity of SCAN and gSCAN. 

In the following sections, we discuss the key components of ReaSCAN. We first introduce the process of generating ReaSCAN commands. Next, we describe how commands are grounded with shape worlds, and specifically the distractor sampling strategies. Finally, we propose test splits which provide systematic tests of a model's generalization abilities.\footnote{We release the version of ReaSCAN used in this paper, and our code to generate ReaSCAN at \url{https://github.com/frankaging/Reason-SCAN}.} We discuss potential ReaSCAN artifacts in \Appref{app:artifacts}.

\begin{figure}[!t]
\begin{minipage}{\textwidth}

\begin{minipage}{0.7\textwidth}
\centering
\resizebox{\linewidth}{!}{%
      \centering
    \begin{tabular}{rll}
      \toprule
      \texttt{Syntax} & \textbf{Descriptions} & \textbf{Expressions} \\ \midrule
      \texttt{\$VV} & verb & \{walk to, push, pull\} \\
      \texttt{\$ADV} & adverb & \{while zigzagging, while spinning, \\
      & & cautiously, hesitantly\} \\
      \texttt{\$SIZE} & attribute & \{small, big\}$^{*}$ \\
      \texttt{\$COLOR} & attribute & \{red, green, blue, yellow\} \\
      \texttt{\$SHAPE} & attribute & \{circle, square, cylinder, box, object\} \\ [1ex]
      \texttt{\$OBJ} & objects & (a | the) \texttt{\$SIZE?} \texttt{\$COLOR?} \texttt{\$SHAPE} \\
      \texttt{\$REL} & relations &  \{in the same row as, in the same column as, \\ && in the same color as,
      in the same shape as, \\ && in the same size as, inside of\} \\ 
      \texttt{\$REL\_CLAUSE} & clause & \texttt{\$REL} \texttt{\$OBJ} \\
      \bottomrule
    \end{tabular}}
\captionof{table}{Definitions of syntax used in ReaSCAN command generation.${}^{*}$the actual size of any shape is chosen from \{1,2,3,4\} as in gSCAN~\cite{ruis2020benchmark}.}\label{tab:ReaSCAN-DSLs}
\end{minipage}
\hfill
\begin{minipage}{0.28\textwidth}
\centering
\captionsetup[subfloat]{labelformat=empty}
\begin{tabular}{cc}
\small same row as &  \small same column as \\
\includegraphics[width = 0.60in]{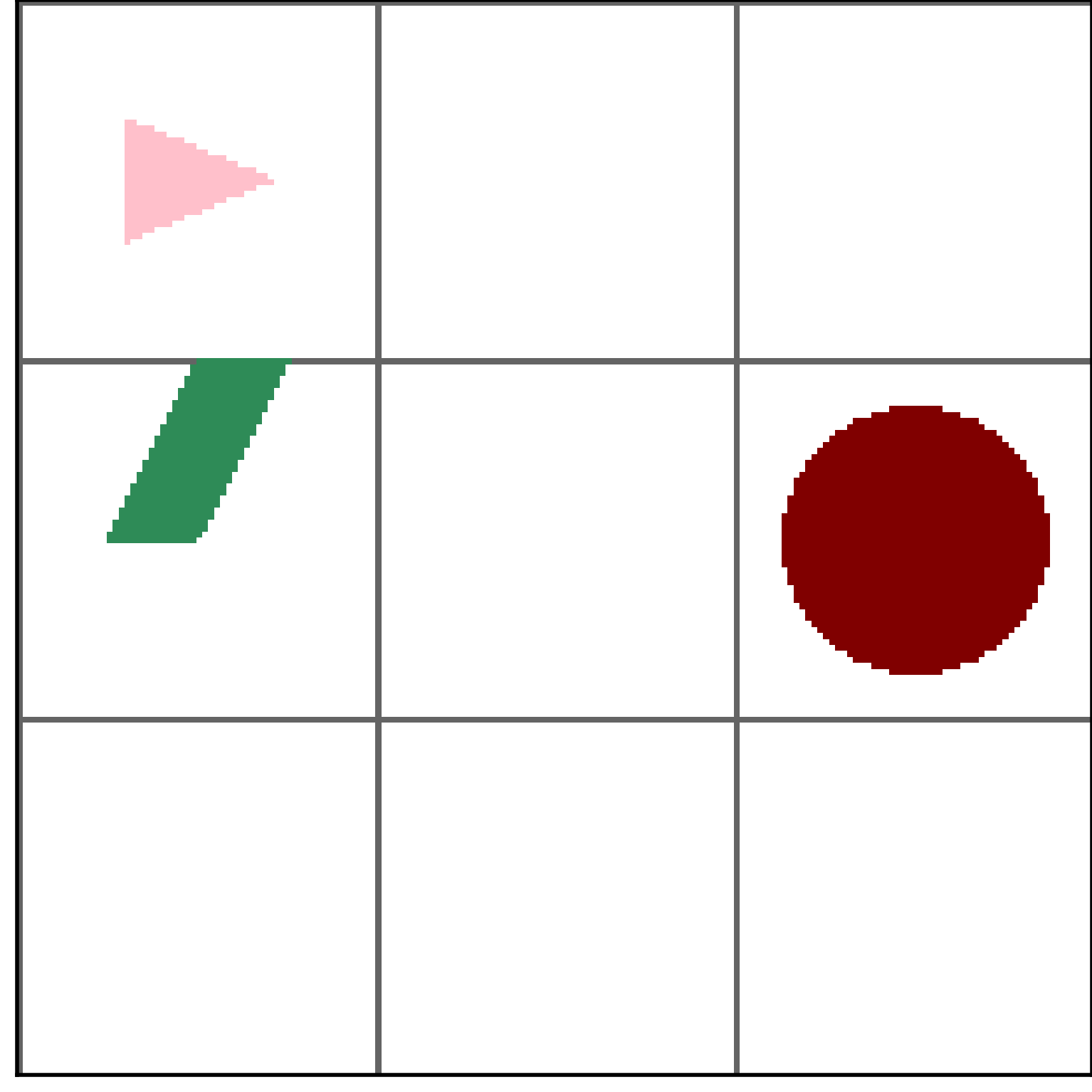}  &
\includegraphics[width = 0.60in]{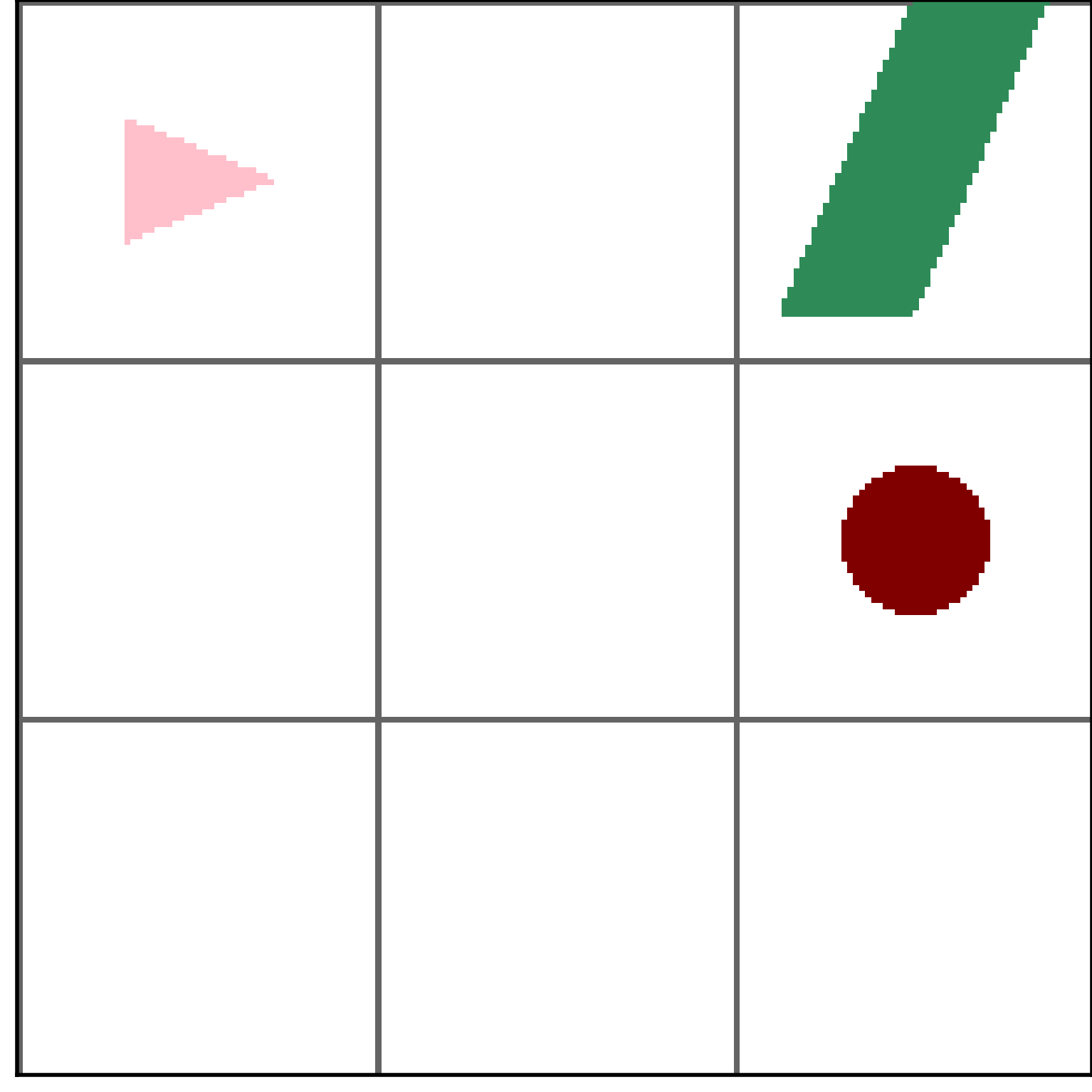} \\
\small same color as &  \small same shape as \\
\includegraphics[width = 0.60in]{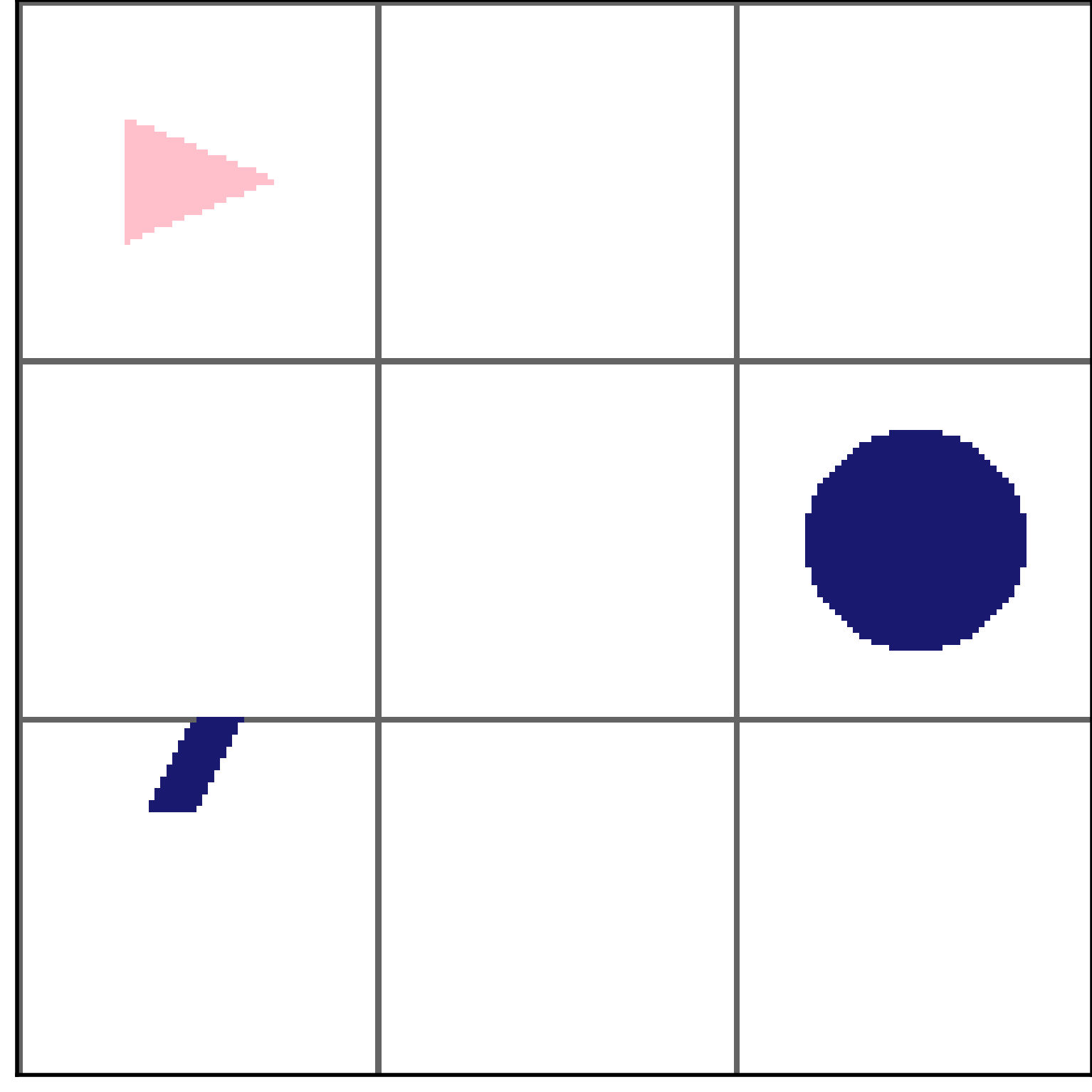} &
\includegraphics[width = 0.60in]{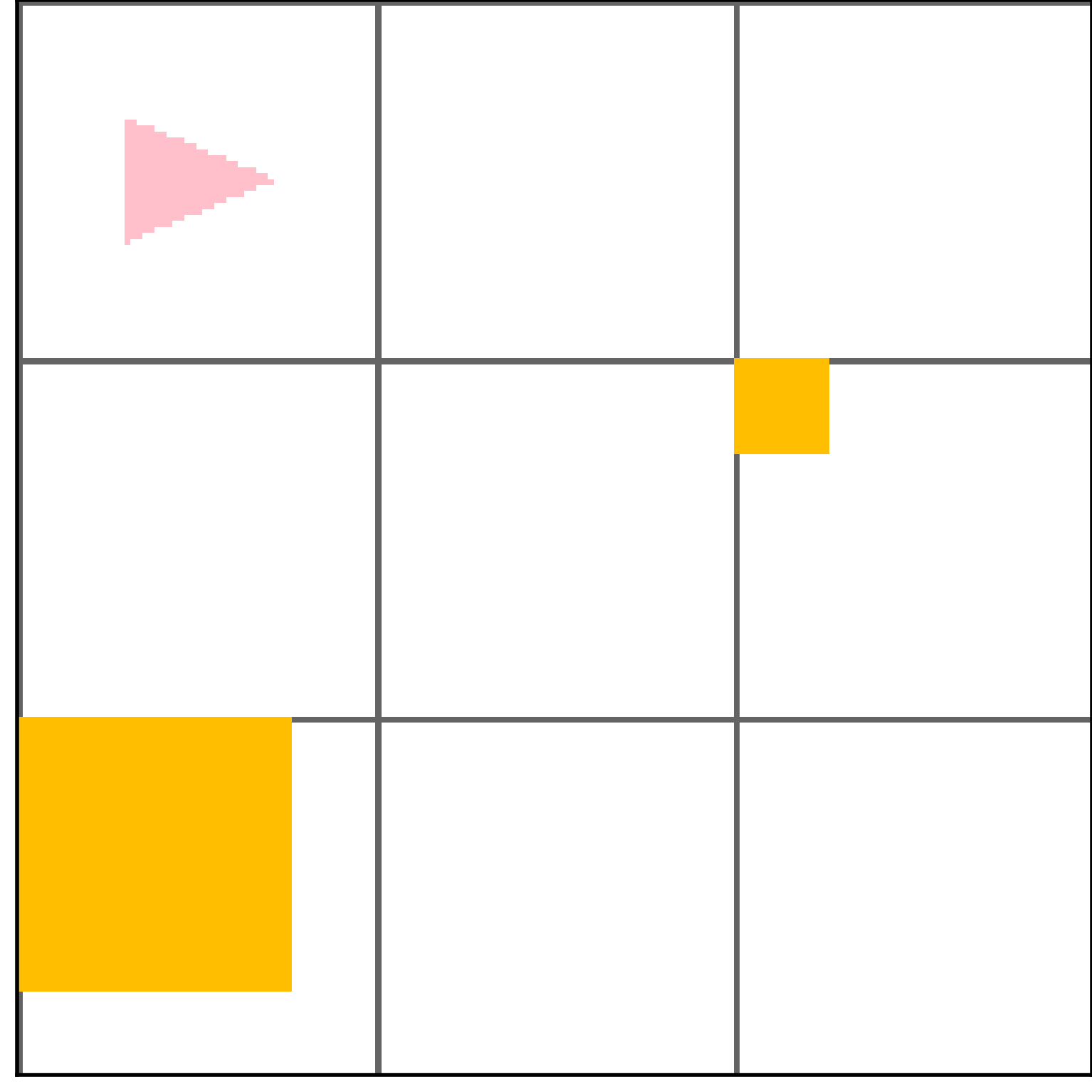} \\
\small same size as &  \small inside of \\
\includegraphics[width = 0.60in]{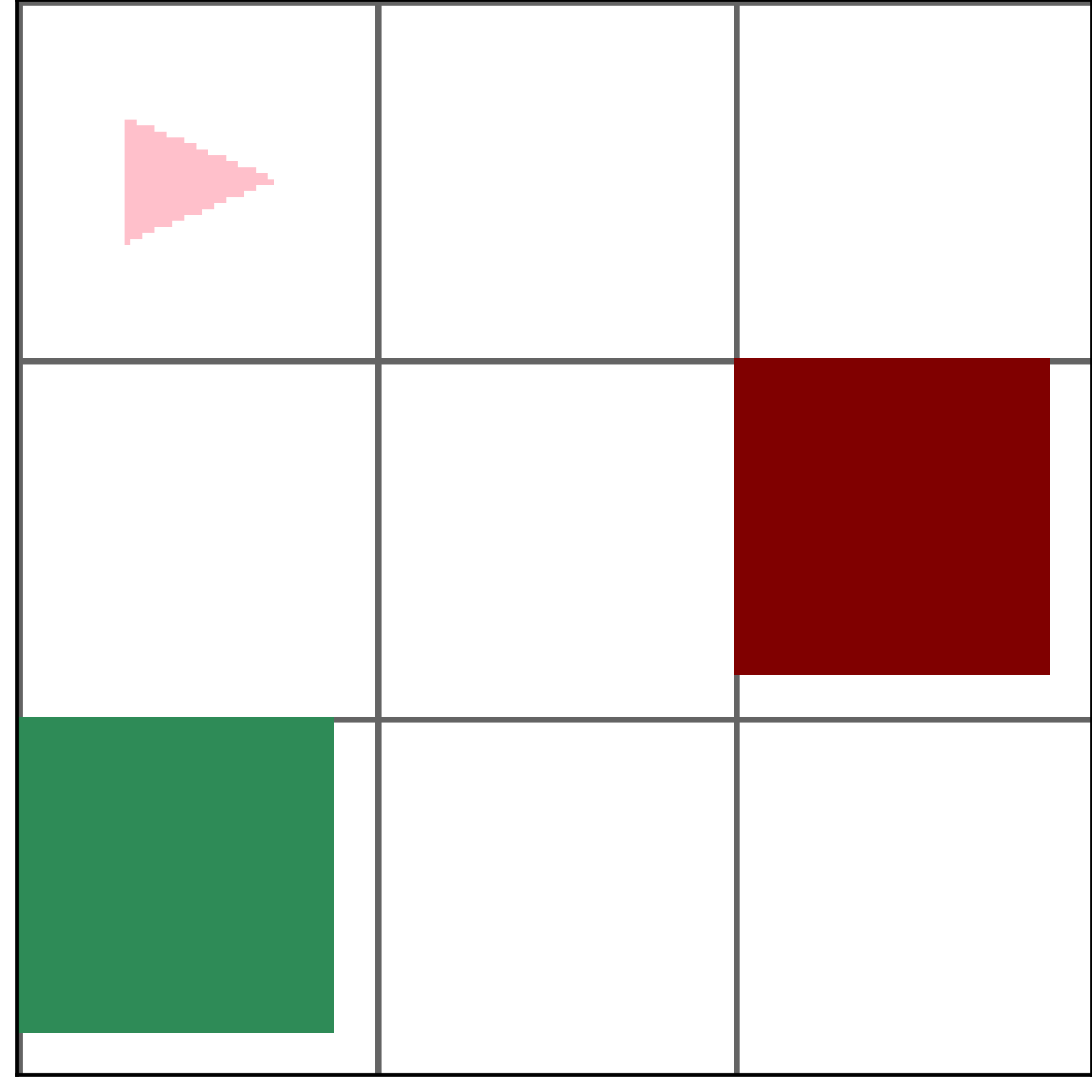} &
\includegraphics[width = 0.60in]{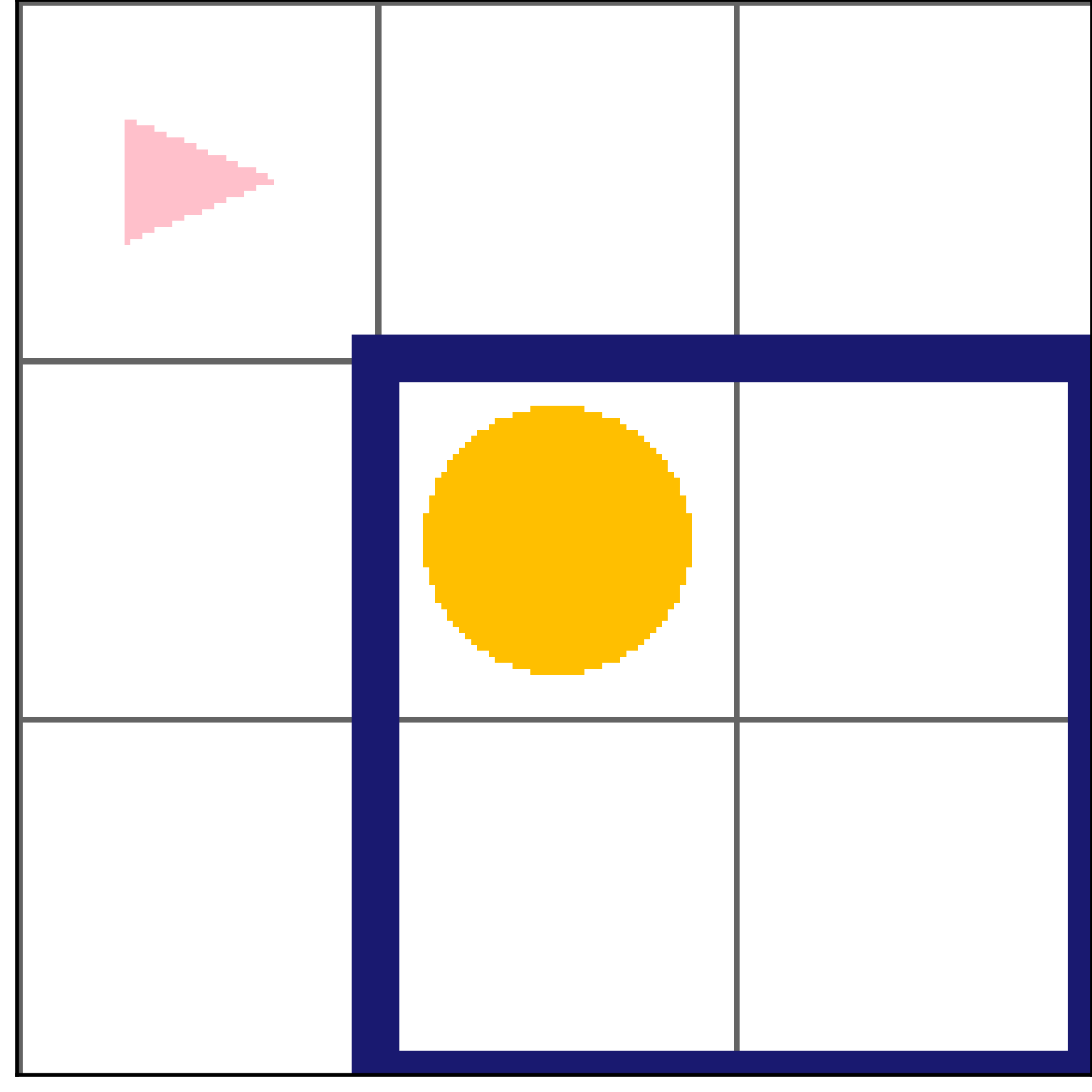} \\
\end{tabular}
\captionof{figure}{Relations.}
\end{minipage}

\end{minipage}
\end{figure}

\subsection{ReaSCAN Command Generation} \label{sec:command-comp}

ReaSCAN commands are constructed with the following regular expression pattern:
\begin{align*}
    \texttt{\text{Pattern}}:= \texttt{\$\text{VV}} \; \texttt{\$\text{OBJ}} \; ( \text{that}\ \text{is}\ \texttt{\$REL\_CLAUSE} \; (\text{and}\  \texttt{\$REL\_CLAUSE})\texttt{*})\texttt{*} \; \texttt{\$\text{ADV}?} \label{eqn:command-pattern}%
\end{align*}
where the recursive structure allows commands to contain multiple relative clauses and conjunctive clauses. If there is no relative clause, the resulting commands are comparable to gSCAN commands (e.g., ``walk to the red square cautiously'').
From the regular expression, commands are created by sampling terms from each class, where classes are indicated by ``\texttt{\$}'' in the pattern as defined in \Tabref{tab:ReaSCAN-DSLs}. 
For example, we substitute $\texttt{\$\text{REL}} \; \texttt{\$\text{OBJ}}$ for $\texttt{\$\text{REL\_CLAUSE}}$, and we can further recursively sample terms from expression classes. 

During this process, we also introduce restrictions to avoid ungrammatical and unnatural commands, enforced by rule-based conditional sampling. This way, commands such as ``walk to the square that is in the same color as the red circle'' would be excluded, as ``walk to the red square'' is a shorter and more direct formulation with the same meaning. See \Appref{app:command} for details about our rule-based conditional sampling over commands.

In this data creation procedure, both the relative clauses and conjunctive clauses have the flexibility to expand in depth and in width. In this paper, we focus on commands with a maximum of a single conjunction of two relative clauses. In total, we generate the following commands:
\begin{itemize}
	\item \texttt{Simple}:= \texttt{\$\text{VV}} \texttt{\$\text{ADV}?} (equivalent to gSCAN commands) 
	\item \texttt{1-relative-clause}:= \texttt{\$\text{VV}} \texttt{\$\text{OBJ}} \text{that}\ \text{is}\ \texttt{\$REL\_CLAUSE} \texttt{\$\text{ADV}?}
	\item \texttt{2-relative-clauses}:= \texttt{\$\text{VV}} \texttt{\$\text{OBJ}} \text{that}\ \text{is}\ \texttt{\$REL\_CLAUSE} \text{and}\  \texttt{\$REL\_CLAUSE} \texttt{\$\text{ADV}?}
\end{itemize}
We use our framework to generate three separate subsets for each command pattern. We then define random train/dev/test splits for each of the subsets to benchmark difficulty (see \Secref{sec:rand-split} for details), where \texttt{Simple} commands are equivalent to gSCAN commands. As shown in \Figref{fig:action-length}, the action sequence length has the same distribution as gSCAN and across all patterns.

\subsection{ReaSCAN World Generation with Active Distractor Sampling} \label{sec:distractor-sampling}

Similar to gSCAN, we use the open-sourced Mini\-Gym from Open-AI\footnote{\url{https://github.com/maximecb/gym-minigrid}} to generate multiple shape worlds for each command. Objects are freely placed in an $n \times n$ grid-world, where we fix $n=6$. Given a command $\textbf{C}_{i}$, objects and their locations are determined as follows:
(1) We select objects mentioned in $\textbf{C}_{i}$, initialize them with their specified features, and randomly fill underspecified features. For instance, in \Figref{fig:graph-conversion}, the command requires the second object to be green and a circle, but its size is not specified and so size is randomly assigned (e.g., here as \emph{big}).
(2) The objects are randomly placed on the grid while ensuring the relations expressed in $\textbf{C}_{i}$ are true.
(3) We sample distractors in a way that ensures that failure to fully understand $\textbf{C}_{i}$ has a high likelihood of leading to an incorrect prediction about the target.

\begin{table}[t!]
\begin{minipage}{0.48\linewidth}
\centering
\includegraphics[width=1\linewidth]{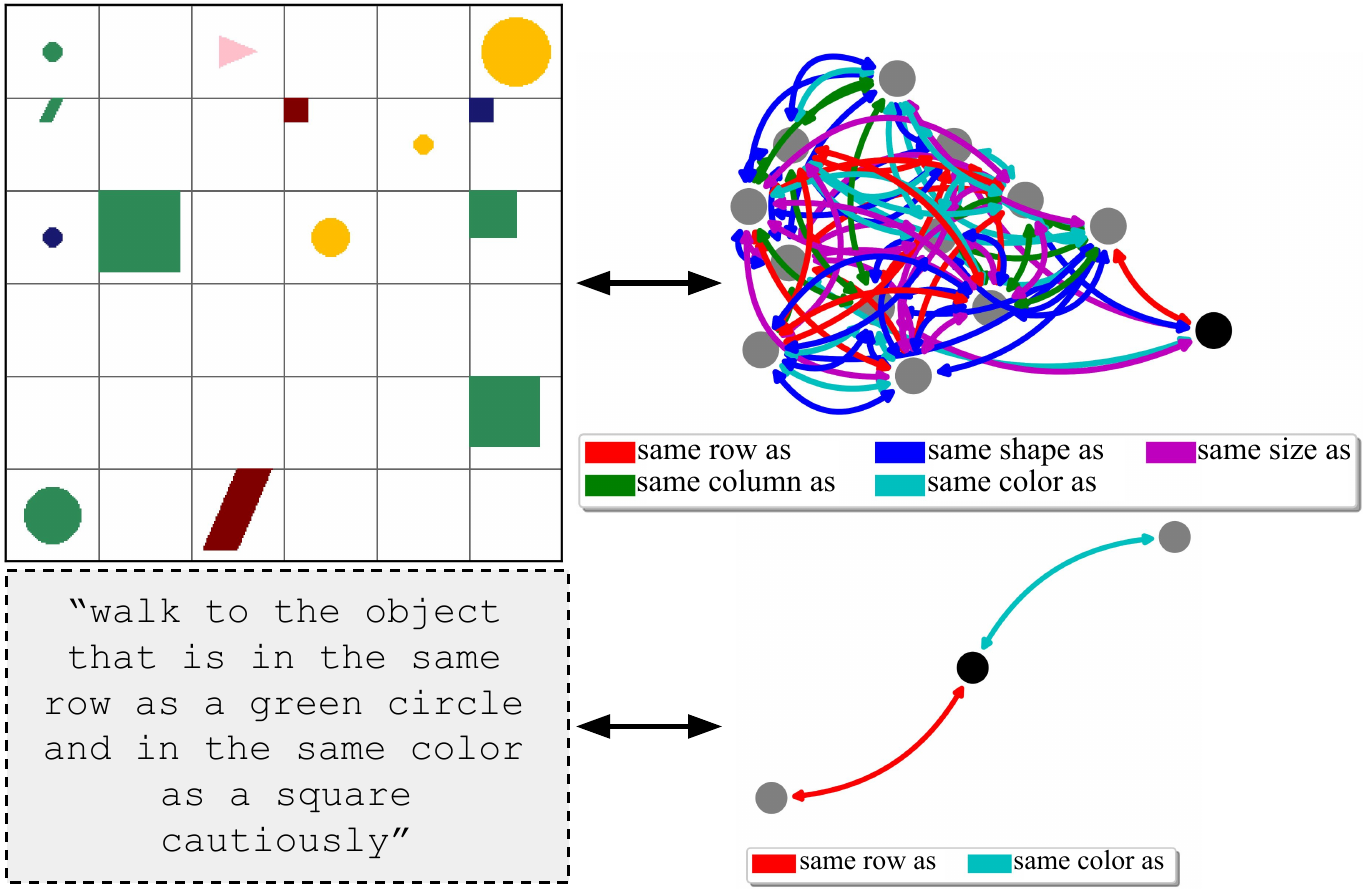}
\captionof{figure}{Illustration of the conversions between the multi-edge graph and the shape world or the command.}
\label{fig:graph-conversion}
\end{minipage}\hfill
\begin{minipage}{0.48\linewidth}
\centering
\includegraphics[width=1\linewidth]{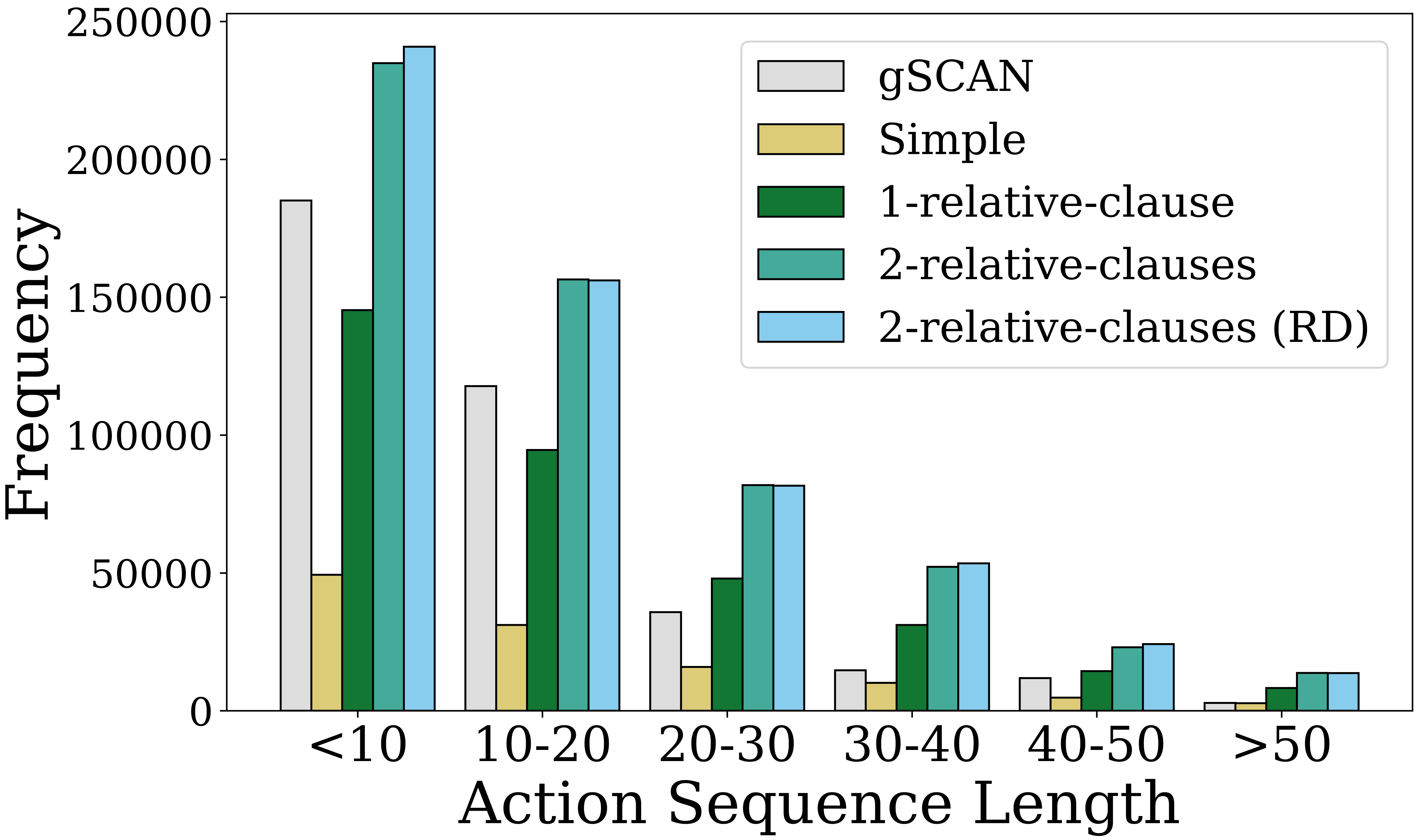}
\captionof{figure}{Length distributions of action sequences for different datasets.}
\label{fig:action-length}
\end{minipage}
\end{table}

As discussed in \secref{sec:gscan}, careful distractor sampling is essential for ensuring that our dataset can be used to assess systems for compositionality. Distractors must reliably introduce uncertainty about the identity of the target.

For example, if the target is a small red circle, a large red circle competes with the target in the size dimension, and confusing the distractor with the target would indicate a lack of understanding of the size domain or its composition. Distractors that have little in common with the target are therefore weak distractors. We employ four distractor-sampling methods that ensure a challenging task that can be used to reliably diagnose specific model shortcomings.
We exemplify their purpose by referring back to the \texttt{2-relative-clauses} example (i.e., the third example) in \Figref{fig:reascan-examples}.

\textbf{Attribute-based distractors} compete with the target if a model struggles with size, color, and shape features. They are created by simulating a change of one of these features in the command and adding objects to the world which make a distractor the plausible target. For instance, if we substitute the shape attribute of the ``blue cylinder'' to ``circle'' in the command, the referent target changes (e.g., it could be object \texttt{D3} in the world). Correctly interpreting the shape attribute becomes crucial for correct target identification.

\textbf{Isomorphism-based distractors} become potential targets after word-order permutations of the command. For instance, if we scramble the word order of the command by swapping attributes between the second and the third objects, and change them to ``small blue circle'' and ``red cylinder'', the referent target changes (e.g., object \texttt{D1} in the world). These distractors are crucial to ensure the necessity of linguistic compositionality to solve the task while Bag-of-Words models can maximally achieve chance accuracy.

\textbf{Relation-based distractors} ensure that the relative clauses in the command are required to identify the intended target referent. For instance, if the model only understands the first relational clause ``the same column as a blue cylinder'', other distractors may become the referent target (e.g., object \texttt{D2-1} in the world); similarly, if the model only understands the second relational clause ``the same row as a small red circle'', object \texttt{D2-2} in the world may become the referent target.

For each world, we sample relation-based distractors exhaustively, and we sample at least one attribute-based distractor by randomly selecting one object and perturbing its attribute. For isomorphism-based distractors, we randomly select any pair of objects and swap attributes if applicable. If a distractor-sampling method cannot work for a specific command-world pair, we incorporate \textbf{random distractors} by randomly sampling a size, color, and shape for each random distractor. This results in a maximum of 16 objects in each generated world. (For gSCAN, the maximum is 12.)

For size, which is described by relative scalar adjectives \citep{KennedyMcNally05}, we added an additional constraint. If the command contains a size modifier, a world always contains a distractor of a different size (similarly to gSCAN). To avoid vagueness about the intended referents, we ensure that there are only two sizes in that particular world. 

As the complexity of distractors increases, there is an increased probability that there could be more than one object in the world that could be the target referent. To ensure a unique solution for all examples, we develop graph-based representations (see \Figref{fig:graph-conversion} for an example) of our shape worlds and use sub-graph matching algorithms to validate examples (see \Appref{app:subgraph-match} for details).

\begin{table}[tp]
    \centering
    
    \resizebox{\linewidth}{!}{%
    \begin{tabular}[c]{l *{3}{r} c *{2}{c} c *{2}{c} c *{2}{c} }
      \toprule
      & \multicolumn{3}{c}{\multirow{2}{*}{Command-World Pairs}} & &  \multicolumn{8}{c}{Exact Match\% (Std.)} \\
      & & & & &  \multicolumn{2}{c}{\texttt{Random}} & &  \multicolumn{2}{c}{\texttt{M-LSTM}} & & \multicolumn{2}{c}{\texttt{GCN-LSTM} } \\
      & Train & Dev & Test & & Dev & Test & & Dev & Test & & Dev & Test \\
      \cmidrule{1-4} \cmidrule{6-7} \cmidrule{9-10} \cmidrule{12-13}
      \texttt{gSCAN}    &  367,933 &  19,282 & 3,716 & & - &  - & &  - &  97.69 (0.22) &  &  - &  98.60 (0.95) \\
      \cmidrule{1-4} \cmidrule{6-7} \cmidrule{9-10} \cmidrule{12-13}
      \texttt{Simple}    &   113,967 &  6,318 &  1,215 & & 0.17 (0.06) &  0.11 (0.13) & &  93.39 (1.97) & 93.64 (2.52) & &  98.06 (0.98) &   97.86 (1.27)\\
      \texttt{1-relative-clause}    &   340,985 &  18,903 & 3,635 & & 0.14 (0.04) &  0.12 (0.02) & &  60.68 (3.04) & 61.28 (1.81) & &  97.25 (0.68) &  97.19 (0.79) \\
      \texttt{2-relative-clauses}    &   549,634 &  30,470 & 5,859 & & 0.12 (0.01) &  0.13 (0.03) & &  53.08 (13.9) & 52.77 (14.6) & &  96.80 (0.82) &   96.85 (0.75) \\
      \texttt{2-relative-clauses} (RD)  &  569,835 &  31,590 & 6,
      075 & & 0.16 (0.02) &  0.12 (0.05) & &  89.56 (0.66) & 89.81 (0.60) & &  98.14 (0.45) &   97.97 (0.48) \\
    \texttt{All} &  539,722 &  29,920 &  5,753 & & 0.13 (0.02) &  0.14 (0.03) & &  78.48 (1.38) & 79.04 (1.24) & &  98.78 (0.55) &  98.96 (0.59) \\
      \bottomrule
    \end{tabular}}
    
    \caption{ReaSCAN statistics with random splits and performance results of baseline models trained separately for each command pattern. \texttt{All} excludes compositional splits. Results are aggregated from 3 independent runs with different random seeds. Performance for gSCAN is from the original papers for \texttt{M-LSTM}~\cite{ruis2020benchmark} and \texttt{GCN-LSTM}~\cite{gao2020systematic}.}
  \label{tab:reascan-data-stats-perf}
\end{table}

\subsection{Compositional Splits} 

ReaSCAN allows us to define a variety of different train/dev/test splits that vary in complexity. \Tabref{tab:comp-split-perf} provides an overview of the splits that we have explored to date. Test splits from category A investigate novel attribute compositions at the command and object level (see \Secref{sec:comp-split-novel-obj-attr}), which are adapted from gSCAN. Test splits in category B investigate how a model generalizes to previously unseen co-occurrences of concepts, including both objects and relations (see \Secref{sec:comp-split-coocurr}), unique to ReaSCAN. Finally, category C investigates if a model can extrapolate from simple to more complex embedded phrase structures (see \Secref{sec:comp-split-rea-gen}).\footnote{We don't report on some splits from gSCAN, such as novel relative agent positions, novel action length, and novel adverbs, since ReaSCAN introduces only minimal changes for these conditions. However, our ReaSCAN pipeline generates these splits as well.}

\section{Models}

We report ReaSCAN experiments with three models. We give high-level descriptions here, and \Appref{app:models} provides additional details.

\paragraph{Random Baseline} A sequence-generation model that randomly samples actions from our vocabulary and generates action sequences with the same lengths as the actual action sequences. This serves as the lower bound of model performance.

\paragraph{M-LSTM} A multimodal LSTM model, which we adapted from a model proposed for gSCAN~\cite{ruis2020benchmark}. %
This is a sequence-to-sequence (seq2seq) model~\cite{sutskever2014sequence} that takes an encoding of the visual input as a separate modality. The encoder consists of two parts: a bidirectional LSTM (BiLSTM; \cite{hochreiter1997long, schuster1997bidirectional}) as the language encoder for the commands, and a convolutional network (CNN)~\cite{fukushima1982neocognitron} as the shape-world encoder. Given a world-command pair $(\mathcal{W}_{i,j}, \textbf{C}_{i})$ as the input, the goal is to generate an action sequence $\textbf{a}_{i,j}$. The output sequence is generated by an attention-based bidirectional LSTM.

\paragraph{GCN-LSTM} A graph convolutional neural (GCN) network with a multimodal LSTM which is, to the best of our knowledge, the currently best-performing model on gSCAN~\cite{gao2020systematic}. The model encodes commands using a BiLSTM with multi-step textual attention~\cite{hudson2018compositional}. The shape world is encoded using a GCN layer. The command embedding is fed into the GCN, which makes it language-conditioned. The nodes in the GCN are initialized with representations of the objects in the shape world, where these representations are binary encodings of the objects' attributes. Then, it performs multi-rounds message passing to contextualize object embeddings based on relations. Then, the object embeddings are fed through another CNN layer before feeding into an attention-based BiLSTM together with the command embedding to generate the output sequence, as in~\citet{ruis2020benchmark}.

\begin{table}[tp]
    \centering
    
 \resizebox{\linewidth}{!}{%
    \begin{tabular}[c]{l c *{3}{c} }
      \toprule
      \multirow{2}{*}{Compositional Splits} & \multirow{2}{*}{Command-World Pairs} & \multicolumn{3}{c}{Exact Match\% (Std.)} \\
      & & \texttt{Random} & \texttt{M-LSTM} & \texttt{GCN-LSTM} \\
      \midrule
      \texttt{Simple (Test)} & 921 & 0.07 (0.06) &  96.27 (0.54) & 99.71 (0.22) \\
      \texttt{1-relative-clause (Test)} & 2,120 & 0.08 (0.07) &  79.09 (2.63) & 99.14 (0.23)  \\
      \texttt{2-relative-clauses  (Test)} & 2,712 & 0.10 (0.02) &  73.16 (1.85) & 98.58 (0.54)  \\
      \texttt{All (Test)} & 5,753 &  0.14 (0.03) & 79.04 (1.24) & 98.96 (0.59) \\
      \midrule
      \texttt{A1:novel color modifier} & 22,057  & 0.12 (0.05) & 50.36 (4.03) & 92.25 (0.77)   \\
      \texttt{A2:novel color attribute} & 81,349  & 0.14 (0.01) &  14.65 (0.55) & 42.05 (4.55) \\
      \texttt{A3:novel size modifier} & 35,675 & 0.14 (0.03)  &  50.98 (3.69) & 87.46 (2.22) \\
      \texttt{B1:novel co-occurrence of objects} & 10,002 & 0.12 (0.03) &  52.17 (1.63) & 69.74 (0.30) \\
      \texttt{B2:novel co-occurrence of relations} & 6,660 & 0.16 (0.05) &  39.41 (1.53) & 52.80 (2.75) \\
      \texttt{C1:novel conjunctive clause length} & 8,375 & 0.10 (0.01) &  49.68 (2.73) & 57.01 (7.99) \\
      \texttt{C2:novel relative clauses} & 8,003 & 0.09 (0.02) &  25.74 (1.36) & 22.07 (2.66) \\
      \bottomrule
    \end{tabular}}
    
    \caption{ReaSCAN statistics with compositional splits and performance results of baseline models trained with all command patterns. Results are aggregated from 3 independent runs with different random seeds.}
  \label{tab:comp-split-perf}
\end{table}

\section{Experiments}

\subsection{Random Split} \label{sec:rand-split} 

We generate large random splits for all patterns to validate that models can learn to follow ReaSCAN commands when there are no systematic differences between training and testing. We do this while systematically varying the complexity of the inputs, from \texttt{Simple} (no relative clauses, as in gSCAN) to \texttt{2-relative-clauses}, and we evaluate when merging all three together (\texttt{All}). \Appref{app:reascan-dataset} provides additional details concerning how these splits are created.

The results in~\Tabref{tab:reascan-data-stats-perf} show that the \texttt{GCN-LSTM} is uniformly superior to the \texttt{M-LSTM}. In addition, for both models, performance drops as the number of relative clauses grows. The \texttt{M-LSTM} performs far worse with longer clauses (43.65\% drop from \texttt{Simple} to \texttt{2-relative-clauses}). The \texttt{GCN-LSTM} experiences smaller drops (1.03\% from \texttt{Simple} to \texttt{2-relative-clauses}). These results suggest that graph-based neural networks may be better at capturing relations between objects and reasoning over relations than the plain CNNs used by the \texttt{M-LSTM}. Additionally, the \texttt{GCN-LSTM} shows smaller standard deviations from random initializations, suggesting it is more robust on the ReaSCAN task as well.

When we resample shape worlds with only random distractors, the performance of both models increases. In fact, with random distractors, test performance of \texttt{2-relative-clauses} drops less than 4\% compared to the \texttt{Simple} conditions, for both models. This finding reinforces the importance of sampling challenging distractors. 

\subsection{A: Novel Object Attributes} \label{sec:comp-split-novel-obj-attr}

Evaluating neural models on unseen combinations of object attributes remains an ongoing challenge~\cite{johnson2017clevr, hill2018understanding, eslami2018neural}. Here, we extend gSCAN's efforts in this area by testing models on unseen composites of size, color, and shape. 

\paragraph{A1: Novel Color Modifier} In this split, we hold out all examples where the commands contain ``yellow square'' for any size (e.g., ``small yellow square'' or ``big yellow square''), meaning that models cannot ground any targets to the expression containing ``yellow square''. However, the train set includes examples with phrases such as ``yellow cylinder'' (52,820 unique examples) and ``blue square'' (90,693 unique examples). At test time, models need to zero-shot generalize in order to interpret ``yellow square'' correctly. Our distractor sampling strategy ensures that the scenario contains relevant non-yellow squares and non-square yellow things, so that both shape and color information needs to be integrated for correct target identification. \Tabref{tab:comp-split-perf} shows that both models perform worse on these splits than with random splits, with the \texttt{M-LSTM} showing the largest drop in performance. While the \texttt{GCN-LSTM} is clearly getting traction on this task, the results show that compositional generalization remains a serious challenge. 

\paragraph{A2: Novel Color Attribute} In this split, we test model performance on a novel combination of the target referent's visual features. To test that, we ensure that red squares are never targets during training. Commands also never contain ``red square'' even in the position of the relations (i.e., inside the relative clauses). However, differently sized red squares are seen during training since they often appear as non-target background objects (266,164 unique examples). We make sure the color attribute is necessary for identifying the target referent, and restrictions apply to objects at all positions in the command. Our results in \Tabref{tab:comp-split-perf} show that this split is slightly harder for both models (with a 81.47\% drop for \texttt{M-LSTM} and a 57.51\% drop for \texttt{GCN-LSTM}) than A1 as models need to learn visual composites of ``red square'' from potential reasoning over background objects. 
Once again, our results suggest that \texttt{GCN-LSTM} is better at generalizing to unseen compositions. 

\paragraph{A3: Novel Size Modifier} Size is a relative concept in our commands; the same object could be a small square in one context and not in another, depending on the sizes of the other squares present. Similar to A1, we evaluate whether models can zero-shot generalize to new size/shape combinations. Specifically, we hold out all commands containing ``small cylinder'', meaning that models have not seen expressions such as ``small cylinder'' or ``small yellow cylinder'' during training. At test time, models need to generalize when a small cylinder in any color is referred to with expressions such as ``small cylinder''. During training, the models still learn the relative meaning of ``small'' by seeing examples containing expressions such as ``small square'' (22,866 unique examples) or ``small red circle'' (23,838 unique examples). In addition to generalizing over new composites, models also cannot simply memorize ``small'' as a specific size (e.g., object of size 2), since the meaning is contextually determined. Similar to A1, we ensure that the size attribute is necessary for identifying the referent target, and restrictions apply to objects at all positions.
\Tabref{tab:comp-split-perf} shows that both models achieve comparable performance to A1, which suggests that the generalization capabilities across unseen color and size composites for both models are similar. \texttt{GCN-LSTM} continues to perform better than \texttt{M-LSTM}, suggesting that it is more successful in generalizing to relative modifiers as well.

\subsection{B: Novel Co-occurrence of Concepts} \label{sec:comp-split-coocurr} 
In this experimental condition, we assess the ability of models to generalize to novel combinations of concepts, including objects and relations at the clause level. 

\paragraph{B1: Novel Co-occurrence of Objects} To construct this split, we first collect all objects (e.g., ``small red circle'' and ``big blue square'') mentioned in the training set. Then, we construct commands with seen objects that never co-occur during training. Additionally, we control commands to only contain co-occurrences of relations that are seen during training. In this condition, the \texttt{GCN-LSTM} continues to outperform the \texttt{M-LSTM} in generalizing to unseen co-occurrences of relations. Compared to novel attribute modifiers (i.e., A1 and A3), \texttt{GCN-LSTM} performance decreases.

\paragraph{B2: Novel Co-occurrence of Relations} In this split, we hold out examples containing commands mentioning both ``same size as'' and ``inside of'' relations, meaning the models never see examples such as ``walk to the object that is the same shape as the red object and that is inside of the red box''. However, in training, models see cases where the relation ``inside of'' co-occurs with other relations, such as ``same row as'' (58,863 unique examples). \Tabref{tab:comp-split-perf} shows that both models perform worse compared to B1. This suggests that generalizing over co-occurrence of relations, which requires novel reasoning about objects, is harder for both model architectures.

\subsection{C: Novel Phrase Structures} \label{sec:comp-split-rea-gen} As shown in \Secref{sec:command-comp}, the number of phrase structures in ReaSCAN can be manipulated. In the following experiments, we test whether a model trained with at most two relative clauses (see \Secref{sec:command-comp} for all patterns) can generalize well to commands with novel phrase structures.

\paragraph{C1: Novel Conjunctive Clause Length} In the first experiment, we generate examples with commands that have one additional conjunction clause (i.e., ``\text{and}\ \texttt{\$REL\_CLAUSE}'' is added to the \texttt{2-relative-clauses} commands). Our results in \Tabref{tab:comp-split-perf} suggest that both models struggle to generalize over longer relative clauses (with a 37.15\% drop for \texttt{M-LSTM} and a 42.39\% drop for \texttt{GCN-LSTM}). Since both models are LSTM-based, it may suggest that LSTM-based models don't generalize well to longer sequences at test time, which has been found in more recent studies~\cite{lake2018generalization}, though some of this may trace to how stop tokens are used \citep{newman-etal-2020-eos}.

\paragraph{C2: Novel Relative Clauses} In this experiment, we generate examples with commands that have two recursive relative clauses (i.e., ``and'' is swapped with ``that is'' in the \texttt{2-relative-clauses} commands)\footnote{We only allow relations to be ``same row as'' and ``same column as'' to avoid invalid commands.}. For this condition, both models result in catastrophic failures (with a 67.43\% drop for \texttt{M-LSTM} and a 77.70\% drop for \texttt{GCN-LSTM}). Our results suggest that GCN is incapable of generalizing over novel recursive relations. The performance degradation of \texttt{GCN-LSTM} may suggest that the fault lies with the way the GCN component embeds relational information in its object representations. This is a strength for known combinations but a potential hindrance for novel ones.

\section{Conclusion}

We introduced the ReaSCAN benchmark, which seeks to build on the insights behind the gSCAN dataset of \citet{ruis2020benchmark} while addressing its shortcomings. ReaSCAN is designed to support controlled assessments of whether models have truly learned grounded, compositional semantics. We find that a state-of-the-art \texttt{GCN-LSTM} model achieves strong results for most of the compositional splits from gSCAN. Results on ReaSCAN, however, suggest that those capabilities are overestimates. Furthermore, ReaSCAN allows for more intricate investigations of the resolution of linguistic structure. The \texttt{GCN-LSTM} model is successful at tasks involving novel linguistic modifiers and novel entity attribute combinations, but it fails to generalize in settings involving novel relation combinations and longer commands. These results indicate that, while we are making progress in achieving grounded, compositional models, many substantial challenges remain.
While ReaSCAN introduces complexity to the problem, via sophisticated distractor sampling strategies and more elaborate input commands, the controlled nature of its input commands means that it is far from tackling the full complexity of natural language. Extending ReaSCAN with interpreted naturalistic English commands would begin to address this limitation.

\section*{Broader Impact}

The ReaSCAN benchmark is designed to facilitate the development of models that can use language in a grounded, compositional fashion. Such research has implications for technology development as well as fundamental research in cognitive science and linguistics. We do not foresee any negative impact on society or on the scientific community stemming directly from this research.

\section*{Acknowledgements}

This research is supported in part by the National Research Foundation, Singapore, under its AI Singapore Program (AISG Award No: AISG2-RP-2020-016), and in part by a Stanford HAI Hoffman--Yee grant.

\small

\bibliography{neurips_2021}
\bibliographystyle{unsrtnat}

\newpage
\appendix

\section*{Appendix for `\PaperTitle'}

\section{Dataset Generation}\label{app:reascan-dataset}

\subsection{Action Sequence} 

Following gSCAN~\cite{ruis2020benchmark}, ReaSCAN agents produce strings composed of action symbols \{\texttt{walk}, \texttt{push}, \texttt{pull}, \texttt{stay}, \texttt{L\_turn}, \texttt{R\_turn}\}. The actions \texttt{push} and \texttt{pull} correspond to the ``push'' and ``pull'' verbs in the command. For the verb ``push'', the agent must push the referent object, where pushing requires moving something as far as possible before hitting a wall or another object. For the verb ``pull'', the agent must pull the referent object, in which case it would pull the object back as far as possible before hitting a wall or another object. Additionally, any object of size 1 or 2 is labeled as \emph{light}, and any object of size 3 or 4 is labeled as \emph{heavy}. If the referent target is a heavy object, the agent needs to push twice or pull twice to move to the next cell (e.g., $\actionSeq{\texttt{push,push}}$ or $\actionSeq{\texttt{pull,pull}}$ ). The optional adverbs at the end of the command may alter action sequences by inserting actions following the adverb. For example, a list of actions $\actionSeq{\texttt{L\_turn,L\_turn,L\_turn,L\_turn}}$ is inserted into the action sequence to fulfill the adverbial ``while spinning''. Since composition generalization on adverbs is not the focus of this paper, we randomly sample adverbs for each command. See the original gSCAN paper for details on adverbs creation~\cite{ruis2020benchmark}.

\subsection{Grounded Determiners} 

ReaSCAN further extends the naturalness of the linguistic input by grounding its determiners. If an NP in the command (e.g., ``red circle'') is preceded by the definite determiner ``the'', there is only one red circle in the world. Otherwise, the object is preceded by the indefinite determiner ``a''.

\subsection{Dataset Statistics} 

Given the rich structure of our commands (see~\Secref{sec:command-comp}), the number of possible commands grows quickly with longer patterns. To ensure we can still generate enough shape worlds per command, we have to down-sample our commands significantly for longer commands. For the \texttt{Simple} command, we exhaustively collect all commands, totalling 675 commands. For commands with one or more relative clauses, we then sample 2,025 commands for \texttt{1-relative-clause} and 3,375 for \texttt{2-relative-clauses}. For each command, we sample 180 shape worlds, which is similar to gSCAN. Our framework is also able to generate the full version of ReaSCAN (i.e., considering all combinations of commands), which, though, uses about 250G of disk space.

\subsection{Infrastructure Setups} 

To generate commands for all three patterns, it takes approximately 16 hours using a single process on a standard CPU cluster. With 50 processes, it takes less than 20 minutes for the largest subset in this paper.

\section{Dataset Artifacts} \label{app:artifacts}

In contrast to realistic datasets, synthetic datasets provide controllable environments for testing a specific aspect of neural models. However, synthetic datasets may produce artifacts induced from the programs generating them. Here, we disclose, as comprehensively as we can, potential artifacts resulting from our data generation process.

\subsection{Non-comprehensive Linguistic Structures}

As discussed in \Secref{sec:command-comp}, commands from ReaSCAN follow a specific linguistic template and are non-comprehensive in covering all linguistic structures. For examples, there is no confusion about where relative clauses attach.
Additionally, the location of occurrence of verbs and adverbs is fixed in our commands. In parallel, we also down-sample our commands for the \texttt{1-relative-clause} and \texttt{2-relative-clauses} conditions, to make ReaSCAN models trainable with reasonable computing resources.

\subsection{Non-comprehensive Distractors} 
To generate a complete list of distractors for commands like ``walk to the red circle that is in the same row as the blue square'', we need to change each attribute (e.g., ``red'', ``circle'', ``blue'' and ``square'') while holding others constant. Consequently, our \texttt{2-relative-clauses} commands could potentially require more than 600 distractors to fulfill a complete sampling of distractors. However, this leads to a dataset that is incomparable to gSCAN, which samples at most 12 distractors. Thus, we randomly select a set of phrases and make them necessary for each command (see \Secref{sec:distractor-sampling} for details about distractor sampling strategies).

We quantify the percentages of different types of distractors appearing in ReaSCAN (see definitions in \Secref{sec:distractor-sampling} and \Figref{fig:artifacts}). Those quantifications are based on the distractors we explicitly generate to fulfill these purposes, therefore serving as a lower-bound estimation for all effective distractors within a world. By chance, some might in fact fulfill multiple purposes, or a random distractor might by chance be highly competitive with the target, resulting in a higher number of effective distractors. As shown in \Figref{fig:artifacts}, relation-based distractors are present in almost all examples with \texttt{1-relative-clause} and \texttt{2-relative-clauses} commands. For \texttt{Simple} and \texttt{1-relative-clause}, we sample attribute-based distractors for almost all examples. For \texttt{2-relative-clauses}, we only sample attribute-based distractors when applicable. For example, we skip sampling attribute-based distractors when there are more than 2 boxes present in the shape world. As a result, attribute-based distractors are present in about 60\% of the examples for \texttt{2-relative-clauses} commands. Random distractors are present in close to 100\% of worlds for simpler commands (e.g., \texttt{Simple} commands), and close to 0\% of worlds for commands with more complex structures (e.g., \texttt{2-relative-clauses} commands).  Additionally, we only sample isomorphism-based distractors when applicable. For example, we only swap attributes between objects that are not the referent target.

\subsection{Shapes and Relations Biases} 

As discussed in \Secref{sec:command-comp} and \Appref{app:command}, we sample commands following a set of rules, which may lead to imbalanced sampling for shapes and relations. For example, ``box'' may only appear after the relational clause ``is inside of''. As a result, the frequencies of ``box'' and the relational clause ``is inside of'' are drastically different from the others. Additionally, we disallow unnatural commands such as ``walk to the red circle that is in the same color as the square'' or ``walk to the circle that is in the same color as the red square'', where the relation is redundant or the unnatural command can be simplified. Following these rules, the frequencies of different relations and attributes could be further stratified. 

\begin{figure}[t!]
\centering
     \includegraphics[width=1.0\textwidth]{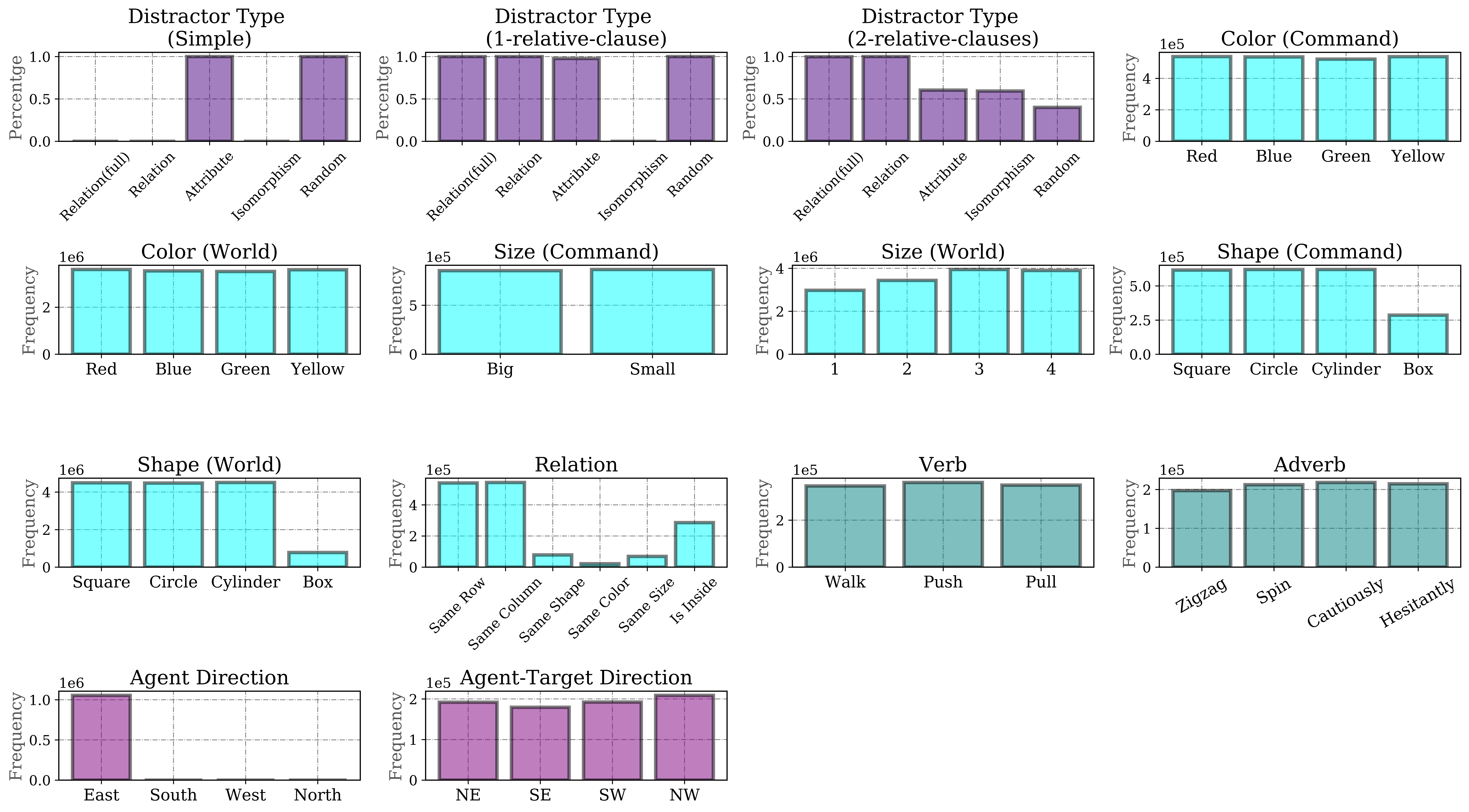}
      \caption{Statistics of ReaSCAN.}
       \label{fig:artifacts}
\end{figure}

\Figref{fig:artifacts} also includes frequency plots for different sizes, colors, shapes, and relations in ReaSCAN. We include frequency distributions for the commands and the shape worlds separately. For colors, frequencies are evenly distributed. For the sizes in commands (e.g., ``big'' or ``small''), frequencies are evenly distributed. For the actual sizes in specific shape worlds, smaller sizes have lower frequencies. This is an artifact due to the shape noun ``box''. Examples including larger boxes may tend to be valid examples compared to smaller ones, which impose spatial limitations. For shapes except the box, frequencies are evenly distributed. ``Box'' is less frequent, as it can only follow the specific relation ``is inside of''. For selected relations, frequencies are extremely biased. As shown in \Figref{fig:artifacts}, relations such as ``same shape as'', ``same color as'', and ``same size as'' are much less frequent than the others. This is due to the fact that we exclude a significant number of unnatural commands that contain these relations (see \Appref{app:command} for details).

\subsection{Self-exclusiveness} 

We assume every object mentioned maps to a unique object in the generated world. For example, if the command is ``walk to the object that is in the same color as the square'', the target is not allowed to be the square itself. This is generally not true in the real world.

\subsection{Other Induced Artifacts}

\Figref{fig:artifacts} includes the distributions for verbs and adverbs present in our commands. As shown in the plot, both verbs and adverbs distribute uniformly. We also include distributions for agent-facing directions and agent--target relative directions at the start. As in gSCAN~\cite{ruis2020benchmark}, our agent always starts facing east. Additionally, the relative directions between the agent and the target at the start are distributed uniformly.

\section{Rule-based Command Samplings}\label{app:command}
As described in \Secref{sec:command-comp}, our command generation process involves building relational clauses between objects. To prevent generated commands from being ungrammatical, we enforce some explicit rules when sampling commands:
\begin{itemize}
	\item \texttt{Rule 1A}: When a ``same shape as'' relational clause is present in the command (e.g., ``\texttt{OBJ1} that is in the same shape as \texttt{OBJ2}''), both objects for this clause cannot contain any shape descriptor. For example, we consider ``a red object that is in the same shape as a red square'' as unnatural since one could just say ``a red square''. If \texttt{OBJ1} contains a shape descriptor already, then the relational clause is unnecessary.
	\item \texttt{Rule 1B}: When the ``same color as'' relational clause is present in the command, both objects for this clause cannot contain any color descriptor. 
	\item \texttt{Rule 1C}: When the ``same size as'' relational clause is present in the command, both objects for this clause cannot contain any size descriptor. For size, we only allow two sizes for any objects when size descriptors are mentioned (e.g., a circle can only be in either of two random sizes if ``small circle'' appears in any command).
	\item \texttt{Rule 2}: The following shape of the ``is inside of'' relational clause must be a box.
	\item \texttt{Rule 3}: For any object term that has relational clauses following it, it cannot be over-specified with descriptors. For example, if we have ``\texttt{OBJ1} that is in the same shape as \texttt{OBJ2}'', we only allow relational conjunction clauses other than ``same shape as''.
	\item \texttt{Rule 4}: In contrast to gSCAN~\cite{ruis2020benchmark}, we enforce an order of modifiers where size descriptors proceed color descriptors (e.g., we allow ``small red circle'' but not ``red small circle'').
\end{itemize}
These rules may over-sample or down-sample certain relations and shapes. We discuss these potential artifacts results in \Secref{app:artifacts}.

\section{Sub-graph Matching} \label{app:subgraph-match}

\subsection{Multi-edge Graph Representation} 
Our distractor sampling strategies incentivize models to learn compositional reasoning. At the same time, these strategies increase the chance that the referent target becomes unidentifiable. For example, even if we randomly place objects in a world, they may form relations consistent with the command by chance. We address this issue by solving constraints using a graph representation.

We represent each world as a graph where nodes represent objects and edges represent relations between objects (see \Figref{fig:graph-conversion} for an example). Each node has attribute-based relations with other nodes. For example, a ``red circle'' node will have a \texttt{SAME\_COLOR} edge to a ``red square'' node. To make sure the referent target is unique for every command--world pair, we ensure only one referent target can be identified by querying the graph with our command. We simplify this problem as a problem of sub-graph matching. We represent each command as a sub-graph where nodes represent objects mentioned in the command and edges encode relations between nodes described in the command. Then, we use a sub-graph matching algorithm~\cite{cordella2004sub} to ensure that the sub-graph representing the command appears only once in the world graph. Sub-graph matching is a NP-hard problem, so we locally optimize our algorithm to have $O(n^k)$ time complexity where $k$ represents the number of clauses in the command. 

\subsection{Locally Optimized Sub-graph Matching Algorithm} 
To ensure that there is only one referent target, we make sure that the sub-graph representing the command only appears once in the graph of the shape world, as illustrated in \Figref{fig:generation-workflow}. 
We include both the complete matching algorithm, which uses \texttt{VF2} as the main algorithm (see \Algref{alg:pattern-match-complete}), and our locally optimized algorithm (see \Algref{alg:pattern-match-local-opti}) for the sub-graph matching problem. Note that this optimized algorithm only applies to three commands mentioned in \Secref{sec:command-comp}, and may need minor modifications to adapt to other commands. We use the \texttt{VF2} algorithm from the \texttt{NetworkX} package for sub-graph matching.\footnote{\url{https://networkx.org/documentation/stable/reference/algorithms/isomorphism.vf2.html}} Full implementations of our algorithms can be found in our code repository.

\begin{algorithm}[ht]
   \caption{\textbf{Complete Multi-edge Sub-graph Matching}}
\label{alg:pattern-match-complete}
\begin{algorithmic}
   \STATE {\bfseries Require} multi-edge directional graphs $\texttt{G}_{w}$ for the world and $\texttt{G}_{c}$ for the command
   \STATE {\bfseries Require} referred object $\texttt{O}$ for the command
   \STATE {\bfseries Return} matching referent targets $\texttt{R}$
\STATE 1: $\texttt{import networkx as nx}$ 
\STATE 1: $\texttt{R} \gets \texttt{\{\}}$ 
\STATE 2: $\texttt{LiG}_{w} \gets \texttt{nx.LineGraph(}\texttt{G}_{w}\texttt{)}$
\STATE 3: $\texttt{LiG}_{c} \gets \texttt{nx.LineGraph(}\texttt{G}_{c}\texttt{)}$
\STATE 4: $\texttt{DiGM} \gets \texttt{nx.VF2(}\texttt{LiG}_{w}, \texttt{LiG}_{c}\texttt{)}$
\STATE 5: \textbf{for} $\texttt{g}_{s} \gets \texttt{DiGM}.\texttt{subgraph\_isomorphisms\_iter()}$ \textbf{do}
\STATE 6: \ \ \ \ \ \  $\texttt{isValid} \gets \texttt{True}$
\STATE 7: \ \ \ \ \ \  \textbf{for} $\texttt{pair}_{w}, \texttt{pair}_{c} \gets \texttt{g}_{s}.\texttt{items}\texttt{()}$ \textbf{do}
\STATE 8: \ \ \ \ \ \ \ \ \ \ \ \ $\texttt{rel}_{w} \gets \texttt{get\_relations(}\texttt{pair}_w\texttt{)}$
\STATE 9: \ \ \ \ \ \ \ \ \ \ \ \ $\texttt{rel}_{c} \gets \texttt{get\_relations(}\texttt{pair}_c\texttt{)}$
\STATE 10: \ \ \ \ \ \ \ \ \ \ \ \ $  \textbf{if not} \; \texttt{rel}_{w} \cap \texttt{rel}_{c}$ \textbf{do}
\STATE 11: \ \ \ \ \ \ \ \ \ \ \ \ \ \ \ \ \ \ $\texttt{isValid} \gets \texttt{False}$
\STATE 12: \ \ \ \ \ \ \ \ \ \ \ \ \ \ \ \ \ \ $  \textbf{break} $
\STATE 13: \ \ \ \ \ \  \textbf{end for}
\STATE 14: \ \ \ \ \ \  \textbf{if} $\texttt{isValid}$ \textbf{do}
\STATE 15: \ \ \ \ \ \ \ \ \ \ \ \ $\texttt{node} \gets \texttt{get\_correspond\_node(O)}$ 
\STATE 16: \ \ \ \ \ \ \ \ \ \ \ \ $\texttt{R} \gets \texttt{R} + \texttt{\{node\}}$ 
\STATE 17: \textbf{end for}
\STATE 18: \textbf{return} $\texttt{R}$ 
\end{algorithmic}
\end{algorithm}

\begin{algorithm}[ht]
   \caption{\textbf{Locally Optimized Multi-edge Sub-graph Matching}}
\label{alg:pattern-match-local-opti}
\begin{algorithmic}
    \STATE {\bfseries Require} multi-edge directional graphs $\texttt{G}_{w}$ for the world and $\texttt{G}_{c}$ for the command
    \STATE {\bfseries Require} referred object $\texttt{O}$ for the command
    \STATE {\bfseries Return} matching referent targets $\texttt{R}$
\STATE 1: $\texttt{R} \gets \texttt{\{\}}$ 
\STATE 2: \textbf{for} $\texttt{n}_w \gets \texttt{G}_{w}.\texttt{get\_nodes()}$ \textbf{do}
\STATE 3: \ \ \ \ \ \ $\texttt{M} \gets \texttt{\{\}}$ 
\STATE 4: \ \ \ \ \ \ $\texttt{rel}_{w} \gets \texttt{n}_w\texttt{.get\_edges()}$
\STATE 5: \ \ \ \ \ \ $\texttt{rel}_{o} \gets \texttt{G}_{c}\texttt{.get\_edges(O)}$
\STATE 6: \ \ \ \ \ \ \textbf{if} $|\texttt{rel}_{o} \cap \texttt{rel}_{w}|\texttt{==}|\texttt{rel}_{o}|$ \textbf{do}
\STATE 7: \ \ \ \ \ \ \ \ \ \ \ \ \texttt{\# found a potential candidate, checking nbrs}
\STATE 8: \ \ \ \ \ \ \ \ \ \ \ \ \textbf{for} $\texttt{nbr} \gets \texttt{n}_w.\texttt{get\_nbrs()}$ \textbf{do}
\STATE 9: \ \ \ \ \ \ \ \ \ \ \ \ \ \ \ \ \ \ \textbf{for} $\texttt{n}_c \gets \texttt{G}_{c}.\texttt{get\_nodes()}$ \textbf{where} $\texttt{n}_c\,\texttt{!=\,O}$ \textbf{do}
\STATE 10: \ \ \ \ \ \ \ \ \ \ \ \ \ \ \ \ \ \ \ \ \ \ \ \ $\texttt{rel}_{\texttt{nbr}} \gets \texttt{\texttt{nbr}}\texttt{.get\_edges()}$
\STATE 11: \ \ \ \ \ \ \ \ \ \ \ \ \ \ \ \ \ \ \ \ \ \ \ \ $\texttt{rel}_{c} \gets \texttt{n}_c\texttt{.get\_edges()}$
\STATE 12: \ \ \ \ \ \ \ \ \ \ \ \ \ \ \ \ \ \ \ \ \ \ \ \ \textbf{if} $|\texttt{rel}_{\texttt{nbr}} \cap \texttt{rel}_{c}|\texttt{==}|\texttt{rel}_{c}|$ \textbf{do}
\STATE 13: \ \ \ \ \ \ \ \ \ \ \ \ \ \ \ \ \ \ \ \ \ \ \ \ \ \ \ \ \ \ \texttt{\# add nbr in to potential matching list}
\STATE 14: \ \ \ \ \ \ \ \ \ \ \ \ \ \ \ \ \ \ \ \ \ \ \ \ \ \ \ \ \ \ $\texttt{M[}\texttt{n}_c\texttt{]} \gets \texttt{M[}\texttt{n}_c\texttt{]} + \texttt{nbr}$
\STATE 15: \ \ \ \ \ \ \ \ \ \ \ \ $\texttt{isValid} \gets \texttt{True}$
\STATE 16: \ \ \ \ \ \ \ \ \ \ \ \ \textbf{for} $\texttt{n}_c \gets \texttt{G}_{c}.\texttt{get\_nodes()}$ \textbf{where} $\texttt{n}_c \texttt{!=O}$ \textbf{do}
\STATE 17: \ \ \ \ \ \ \ \ \ \ \ \ \ \ \ \ \ \ \textbf{if} $|\texttt{M[}\texttt{n}_c\texttt{]}| \texttt{==}0$ \textbf{do}
\STATE 18: \ \ \ \ \ \ \ \ \ \ \ \ \ \ \ \ \ \ \ \ \ \ \ \ $\texttt{isValid} \gets \texttt{False}$
\STATE 19: \ \ \ \ \ \ \ \ \ \ \ \ \ \ \ \ \ \ \ \ \ \ \ \ \textbf{break}
\STATE 20: \ \ \ \ \ \ \ \ \ \ \ \ \textbf{if}  $\texttt{isValid} \texttt{ \&\& } \texttt{at\_least\_one\_unique\_for\_each(\texttt{M})}$ \textbf{do} 
\STATE 21: \ \ \ \ \ \ \ \ \ \ \ \ \ \ \ \ \ \ $\texttt{R} \gets \texttt{R} + \texttt{\{n}_w\texttt{\}}$ 
\STATE 22: \textbf{end for}
\STATE 23: \textbf{return} $\texttt{R}$ 
\end{algorithmic}
\end{algorithm}

\section{Models and Experiments Setups}\label{app:models}

For our \texttt{M-LSTM}\footnote{ \url{https://github.com/LauraRuis/multimodal_seq2seq_gSCAN}.} and \texttt{GCN-LSTM}\footnote{ \url{https://github.com/HQ01/gSCAN_with_language_conditioned_embedding}.} models, we adapt code from the original repositories. For both models, we optimize for cross-entropy loss using Adam with default parameters~\cite{kingma2015adam}. 
The learning rate starts at $1e^{-4}$ and decays by $0.9$ every 20,000 steps for the \texttt{M-LSTM} model. The learning rate starts at $8e^{-4}$ for the \texttt{GCN-LSTM} model with the same learning rate decaying schedule. We train for 200,000 steps, with batch size 2000 for the \texttt{M-LSTM} model, and train for 100 epochs with batch size 200 for the \texttt{GCN-LSTM} model. We choose the best model during training by the performance on a smaller development set of 2,000 examples, which is consistent with the training pipeline proposed in \citet{ruis2020benchmark} for gSCAN. For \texttt{M-LSTM}, we choose the kernel size for the CNN to be 7. For \texttt{GCN-LSTM}, we choose the number of message passing iterations to be 4. We adapt the code released by each paper, which only supports single-GPU training. The training time is about 3 days on a Standard GeForce RTX 2080 Ti GPU with 11GB memory. To foster reproducibility, we release our adapted evaluation scripts in our code repository.

In generating random splits (\secref{sec:rand-split}), we randomly partition train/dev/test after command--world pairs are generated. As shown in \Tabref{tab:reascan-data-stats-perf}, we generate more than 1M example--world pairs in total. Our \texttt{Simple} set is comparable to \texttt{gSCAN}~\cite{ruis2020benchmark}. However, our \texttt{Simple} set is smaller in size since \texttt{gSCAN} permutes based on the agent's relative direction against the referent target. Note that for \texttt{1-relative-clause} and \texttt{2-relative-clauses}, we down-sample our commands since we keep world per command approximately the same as \texttt{gSCAN} to ensure a fair comparison. See \Appref{app:reascan-dataset} for detailed dataset statistics. To evaluate our distractor sampling strategies, we regenerate a new dataset for \texttt{2-relative-clauses} containing only random distractors and analyze model performance results. Finally, we combine all three subsets for all patterns and evaluate aggregated performance (see \Tabref{tab:comp-split-perf} for per pattern performance).

\section{ReaSCAN Generation Workflow}
\begin{figure}[t!]
\centering
     \includegraphics[width=1.0\textwidth]{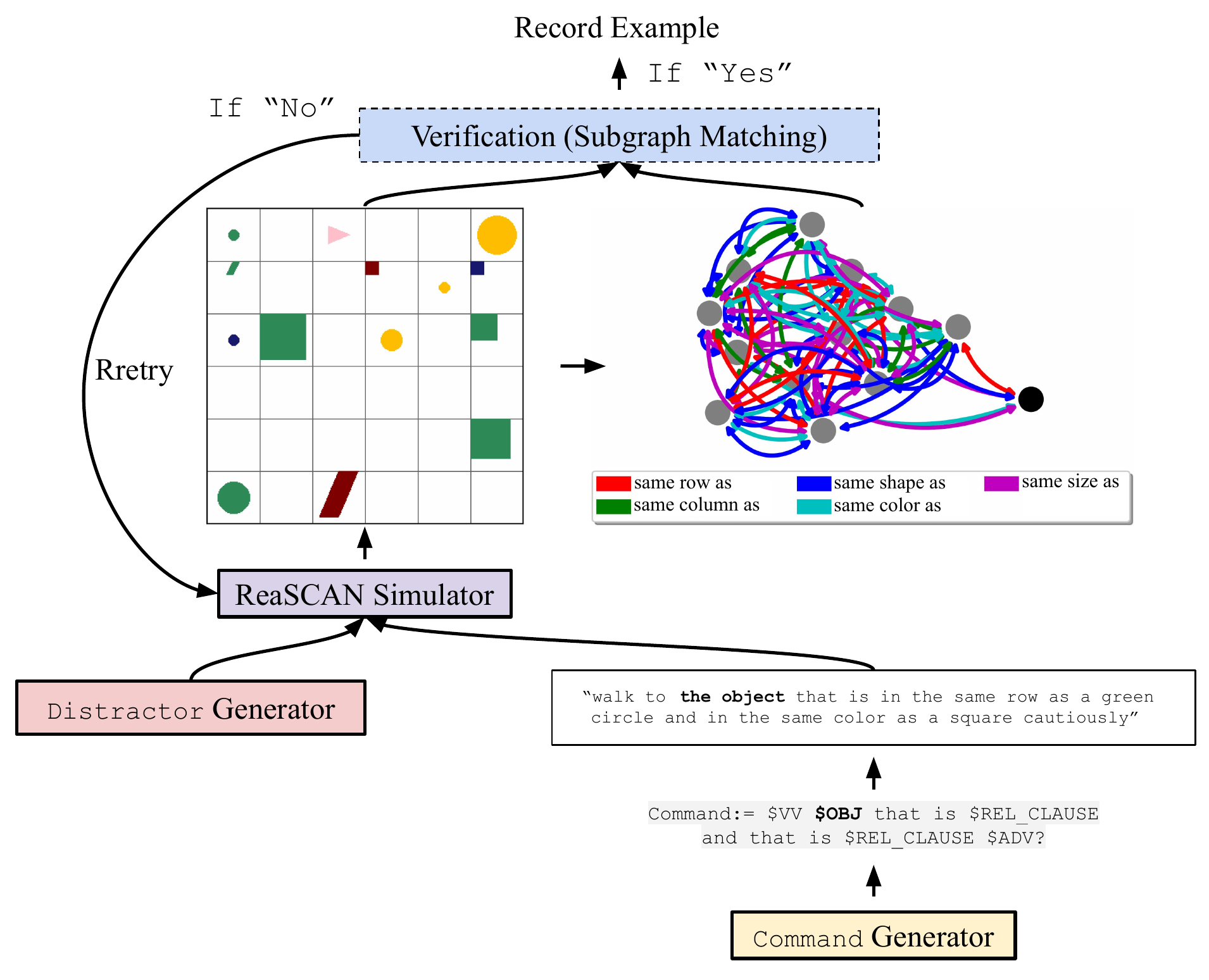}
      \caption{Data generation workflow with a simplified example.}
       \label{fig:generation-workflow}
\end{figure}

\Figref{fig:generation-workflow} illustrates our main data generation workflow with an example with a single relational clause. The data generation workflow contains five main steps:
\begin{itemize}
	\item \texttt{Step 1}: We first sample a command pattern from the \texttt{command} generator. The command pattern contains the basic structure of the command as in~\Secref{sec:command-comp}.
	\item \texttt{Step 2}: We fill out the command pattern by supplying its semantics and generate a fully-formed command.
	\item \texttt{Step 3}: Using our simulator, we generate a shape world containing objects conforming to our command as well as distractors. We have four types of distractors, as described in \Secref{sec:distractor-sampling}. 
	\item \texttt{Step 4}: We then build a graph describing objects and their relations in the shape world generated from the previous step. Using our sub-graph matching algorithm (\Appref{app:subgraph-match}), we validate whether there is only one referent target presented in the shape world grounding the command.
	\item \texttt{Step 5}: If ``yes'', we record the command--world pair. If ``no'', we fall back to \texttt{Step 3} and generate a new shape world.
\end{itemize}

\section{ReaSCAN Examples}\label{app:reascan-examples}
\begin{figure}[t!]
\centering
     \includegraphics[width=0.9\textwidth]{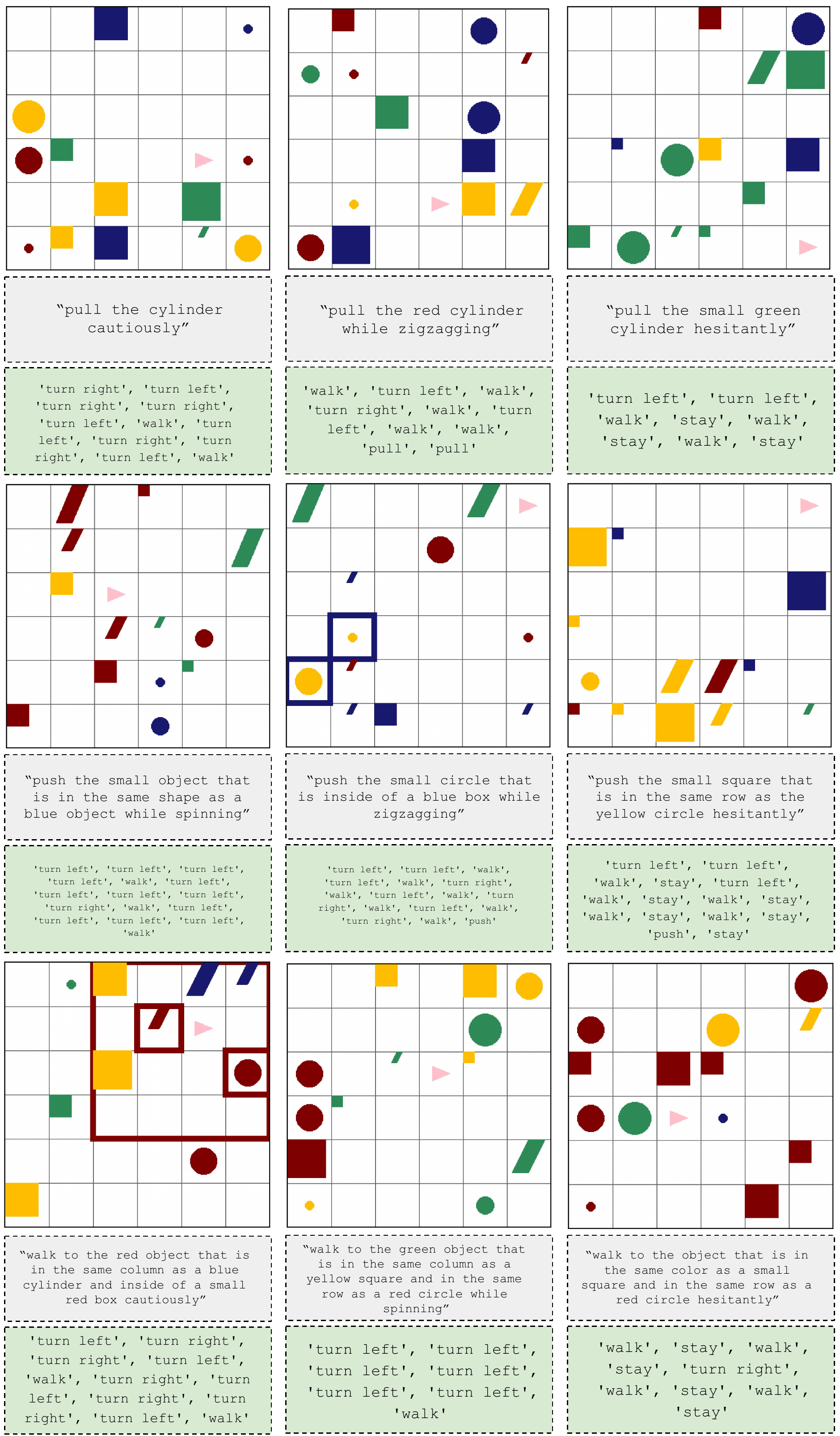}
      \caption{ReaSCAN examples with varying command patterns. The navigation commands and the target action sequences are in the grey boxes and green boxes respectively.}
       \label{fig:more-examples}
\end{figure}

\Figref{fig:more-examples} shows more examples for different patterns of commands in ReaSCAN.

\section{ReaSCAN as an Abstract Reasoning Challenge}\label{app:reascan-arc}

\begin{figure}[t!]
\centering
     \includegraphics[width=1.0\textwidth]{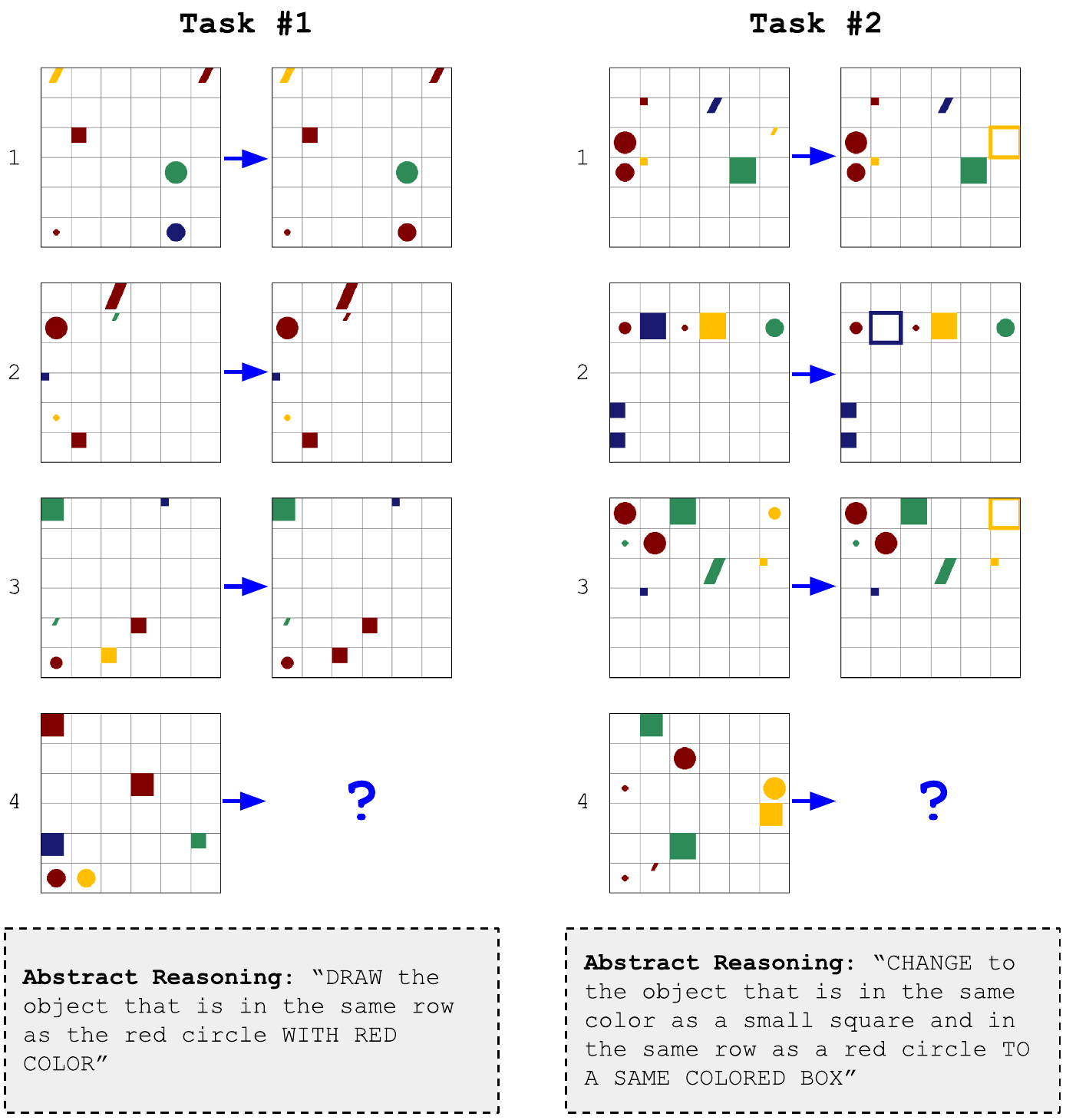}
      \caption{Two simplified abstract reasoning challenges with ReaSCAN. The task mimics human reasoning tests: given a set of input--output (input on the left and output on the right) pairs, the task taker needs to guess the output for the last input. For each task, we provide one potential abstract reasoning statement to solve the task.}
       \label{fig:reascan-arc}
\end{figure}

\Figref{fig:reascan-arc} shows two simplified examples of how we can transform ReaSCAN into an abstract reasoning challenge following the Abstract Reasoning Corpus proposed by~\citet{chollet2019measure}. Instead of generating the action sequence based on a command--world pair, the task is now defined as predicting the output of a test input world, given multiple pairs of input--output worlds as training examples. This can be seen as a program induction or program synthesis task as well. Our pipeline is able to generate the reasoning logic involved in each task with natural language. This task mimics abstract reasoning tests for humans. To generate reasoning tasks, we first extend our current framework to generate multiple shape worlds for each command along with the position of the referent targets for each world. Then, we define some primitive actions (e.g., \texttt{DRAW} for changing color; \texttt{CHANGE} for changing shape) that can be operated on the referent targets. Finally, we generate a new set of shape worlds with operated referent targets.

\section{Dataset Documentation}\label{app:reascan-document}
We have made our dataset and the framework to generate the dataset publicly avaliable at \url{https://github.com/frankaging/Reason-SCAN}. We bear all responsibility in case of violation of rights. The dataset is released with Creative Commons Attribution 4.0 International License. Updates will be reflected in our code repository. The first release is versioned as ReaSCANv1.0. Any subsequent releases will have higher version numbers.

The dataset and its metadata can be found in the code repository. Additionally, we provide detailed steps for how to regenerate ReaSCAN in our code repository. Since the data generation framework is novel, it is self-contained as well. The dataset and its code repository will remain publicly available. We include our datasheets.
\clearpage

\newpage
\appendix

\section*{Datasheet for `\PaperTitle'}
\label{datasheets}

\section{Motivation}

\subsection{For what purpose was the dataset created? Was there a specific task in mind? Was there a specific gap that needed to be filled? Please provide a description}

The dataset was created as a new benchmark for evaluating compositional generalization of neural models by addressing the limitations of gSCAN. We find that gSCAN has three major limitations: (1) its set of instructions is so constrained that compositional interpretation is not required; (2) the distractor objects in its grounded scenarios are mostly not relevant for accurate language understanding; and (3) in many examples, not all modifiers in the command are required for successful navigation, which further erodes the need for compositional interpretation and inflates model performance scores. 

\subsection{Who created this dataset (e.g., which team, research group) and on behalf of which entity (e.g., company, institution, organization)?}
The dataset was created by Zhengxuan Wu, Elisa Kreiss, and Christopher Potts at Stanford University, and Desmond C.~Ong at National University of Singapore.

\subsection{Who funded the creation of the dataset? If there is an associated grant, please provide the name of the grantor and the grant name and number}

This research is supported in part by the National Research Foundation, Singapore, under its AI Singapore Program (AISG Award No: AISG2-RP-2020-016), and in part by a Stanford HAI Hoffman--Yee grant.

\section{Composition}

\subsection{What do the instances that comprise the dataset represent (e.g., documents, photos, people, countries)? Are there multiple types of instances (e.g., movies, users, and ratings; people and interactions between
them; nodes and edges)? Please provide a description.} \label{datasheets:instance}
This dataset is a synthetic dataset. Each instance contains: 
\begin{enumerate}
\item A command--world pair where the command is a synthetic English sentence and the world is a synthetic $n \times n$ grid-world, where we fix $n=6$, using the open-sourced Mini\-Gym from Open-AI.\footnote{\url{https://github.com/maximecb/gym-minigrid}} The world is represented by a list of objects that are present in the world. Each object is defined by a unique tensor.
\item An agent initial position and facing direction.
\item The position of the referent target.
\item The gold label, which is the correct action sequence the agent can execute to reach and operate on the referent target.
\end{enumerate}

\subsection{Does the dataset contain all possible instances or is it a sample (not
necessarily random) of instances from a larger set? If the dataset is
a sample, then what is the larger set? Is the sample representative of the
larger set (e.g., geographic coverage)? If so, please describe how this
representativeness was validated/verified. If it is not representative of the
larger set, please describe why not (e.g., to cover a more diverse range of
instances, because instances were withheld or unavailable).}

ReaSCANv1.0 is provided in our project Github  repository.\footnote{\url{https://github.com/frankaging/Reason-SCAN}} It is sampled from a larger set. As our dataset provides a general framework that can scale up to many more examples, it reaches a point where regular computing resources might not sufficient. %
As a result, we randomly sample a subset from our larger pool. We document known potential artifacts from our generation process in \Appref{app:artifacts}.

\subsection{What data does each instance consist of? “Raw” data (e.g., unprocessed text or images)or features? In either case, please provide a description}
We provide details about our instances in~\Secref{datasheets:instance}.

\subsection{Is there a label or target associated with each instance? If so, please provide a description.}

The gold label is the correct action sequence. The agent can execute the action sequence to reach and operate on the referent target.

\subsection{Is any information missing from individual instances? If so, please
provide a description, explaining why this information is missing (e.g., because it was unavailable). This does not include intentionally removed
information, but might include, e.g., redacted text.}
Everything is included. No data is missing.

\subsection{Are relationships between individual instances made explicit (e.g.,
users’ movie ratings, social network links)? If so, please describe
how these relationships are made explicit.}
N/A.

\subsection{Are there recommended data splits (e.g., training, development/validation, testing)? If so, please provide a description of these
splits, explaining the rationale behind them.}
As our dataset is designed for compositional generalization, we provide train/dev/test splits as well as compositional splits in our released dataset. We also provide scripts to generate these splits in our code repository.

\subsection{Are there any errors, sources of noise, or redundancies in the dataset? If so, please provide a description.}
N/A.

\subsection{Is the dataset self-contained, or does it link to or otherwise rely on
external resources (e.g., websites, tweets, other datasets)?}
Yes, the dataset is self-contained.

\subsection{Does the dataset contain data that might be considered confidential
(e.g., data that is protected by legal privilege or by doctorpatient confidentiality, data that includes the content of individuals non-public
communications)? If so, please provide a description.}
No, this is a synthetic dataset.

\subsection{Does the dataset contain data that, if viewed directly, might be offensive, insulting, threatening, or might otherwise cause anxiety? If so,
please describe why.}
No, this is a synthetic dataset containing synthetic navigation instructions in English and synthetic grid worlds.

\subsection{Does the dataset relate to people? If not, you may skip the remaining
questions in this section.}
No, this is a synthetic dataset and does not contain any human-assigned labels.

\subsection{Does the dataset identify any subpopulations (e.g., by age, gender)?
If so, please describe how these subpopulations are identified and provide
a description of their respective distributions within the dataset.}
N/A.

\subsection{Is it possible to identify individuals (i.e., one or more natural persons), either directly or indirectly (i.e., in combination with other
data) from the dataset? If so, please describe how.}
N/A.

\subsection{Does the dataset contain data that might be considered sensitive in
any way (e.g., data that reveals racial or ethnic origins, sexual orientations, religious beliefs, political opinions or union memberships, or
locations; financial or health data; biometric or genetic data; forms of
government identification, such as social security numbers; criminal
history)?}
N/A.

\section{Collection Process}
N/A. This is a synthetic dataset containing synthetic navigation instructions in English and synthetic grid worlds. As a result, we do not collect any human data.

\section{Preprocessing/cleaning/labeling}

\subsection{Was any preprocessing/cleaning/labeling of the data done (e.g., discretization or bucketing, tokenization, part-of-speech tagging, SIFT
feature extraction, removal of instances, processing of missing values)? If so, please provide a description. If not, you may skip the remainder of the questions in this section.}
No, the synthetic dataset is provided as-is.

\subsection{Was the “raw” data saved in addition to the preprocessed/cleaned/labeled data (e.g., to support unanticipated
future uses)? If so, please provide a link or other access point to the
“raw” data.}
N/A.

\subsection{Is the software used to preprocess/clean/label the instances available? If so, please provide a link or other access point}
N/A.

\section{Use}

\subsection{Has the dataset been used for any tasks already? If so, please provide
a description.}
No, this is our first release of the dataset.

\subsection{Is there a repository that links to any or all papers or systems that
use the dataset? If so, please provide a link or other access point.}
No, this is our first release of the dataset.

\subsection{What (other) tasks could the dataset be used for?}
This dataset is designed for evaluating compositional generalization of neural models as a synthetic navigation task.  This task can be used for referring expression resolution as well.

\subsection{Is there anything about the composition of the dataset or the way it
was collected and preprocessed/cleaned/labeled that might impact
future uses? For example, is there anything that a future user might need
to know to avoid uses that could result in unfair treatment of individuals or
groups (e.g., stereotyping, quality of service issues) or other undesirable
harms (e.g., financial harms, legal risks) If so, please provide a description. Is there anything a future user could do to mitigate these undesirable
harms?}
There is minimal risk for harm: this is a synthetic dataset and does not contain any human labels.

\subsection{Are there tasks for which the dataset should not be used? If so, please
provide a description.}
No, this dataset is used for training neural models that solve the synthetic task posed by the dataset only. The dataset should not be used directly in any real-world applications.

\section{Distribution}

\subsection{Will the dataset be distributed to third parties outside of the entity (e.g., company, institution, organization) on behalf of which the
dataset was created? If so, please provide a description.}

Yes, the dataset is publicly available on the internet.

\subsection{How will the dataset will be distributed (e.g., tarball on website, API,
GitHub)? Does the dataset have a digital object identifier (DOI)?}

The dataset is publicly avaliable at our Github repository.

\subsection{When will the dataset be distributed?}
The dataset is first released in 2021.

\subsection{Will the dataset be distributed under a copyright or other intellectual
property (IP) license, and/or under applicable terms of use (ToU)? If
so, please describe this license and/or ToU, and provide a link or other
access point to, or otherwise reproduce, any relevant licensing terms or
ToU, as well as any fees associated with these restrictions.}
Our dataset has a Creative Commons Attribution 4.0 International License. The dataset is publicly available on the internet. People are allowed to use our scripts to generate their own version of ReaSCAN as well.

\subsection{Do any export controls or other regulatory restrictions apply to the
dataset or to individual instances? If so, please describe these restrictions, and provide a link or other access point to, or otherwise reproduce,
any supporting documentation.}
Unknown.

\section{Maintenance}

\subsection{Who is supporting/hosting/maintaining the dataset?}
Zhengxuan Wu is supporting/maintaining the dataset.

\subsection{How can the owner/curator/manager of the dataset be contacted
(e.g., email address)?}
  \texttt{\href{mailto:wuzhengx@stanford.edu}{wuzhengx@stanford.edu}}.

\subsection{Is there an erratum? If so, please provide a link or other access point.}
There is not an explicit erratum, but updates and fixes with the dataset will be reflected in our Github repository. The first release is versioned as ReaSCANv1.0. Any subsequent releases will have higher version numbers.

\subsection{Will the dataset be updated (e.g., to correct labeling errors, add
new instances, delete instances)? If so, please describe how often, by
whom, and how updates will be communicated to users (e.g., mailing list,
GitHub)?}
This will be posted on the dataset Github repository.

\subsection{If the dataset relates to people, are there applicable limits on the retention of the data associated with the instances (e.g., were individuals in question told that their data would be retained for a fixed period
of time and then deleted)? If so, please describe these limits and explain
how they will be enforced.}
N/A.

\subsection{Will older versions of the dataset continue to be supported/hosted/maintained? If so, please describe how. If not,
please describe how its obsolescence will be communicated to users.}
N/A.

\subsection{If others want to extend/augment/build on/contribute to the dataset,
is there a mechanism for them to do so? If so, please provide a description. Will these contributions be validated/verified? If so, please describe
how. If not, why not? Is there a process for communicating/distributing
these contributions to other users? If so, please provide a description.}
Others may do so and should contact the original authors about incorporating fixes/extensions.

\end{document}